\documentclass{article}

\usepackage[utf8]{inputenc} 
\usepackage[T1]{fontenc}    
\usepackage[]{hyperref}       
\usepackage{url}            
\usepackage{booktabs}       
\usepackage{nicefrac}       
\usepackage{microtype}      

\usepackage{graphicx}
\usepackage{amsmath,amsfonts,amssymb,amsthm}
\usepackage{float}
\usepackage[]{hyperref}
\usepackage[mathscr]{euscript}
\usepackage[dvipsnames]{xcolor}
\usepackage{graphicx}
\usepackage{subcaption}
\usepackage{multicol}
\usepackage{natbib}

\usepackage[accepted]{icml2020}

\newcommand{\ubf}{\mathbf{u}}
\newcommand{\vbf}{\mathbf{v}}

\newcommand{\xbf}{\mathbf{x}}

\newcommand{\Ebf}{\mathbf{E}}

\newcommand{\Acal}{\mathcal{A}}
\newcommand{\Bcal}{\mathcal{B}}

\newcommand{\Ncal}{\mathcal{N}}

\newcommand{\Scal}{\mathcal{S}}
\newcommand{\Tcal}{\mathcal{T}}

\newcommand{\Zcal}{\mathcal{Z}}

\newcommand{\Pscr}{\mathscr{P}}

\newcommand{\Wscr}{\mathscr{W}}

\DeclareMathOperator*{\argmin}{arg\,min}
\DeclareMathOperator*{\argmax}{arg\,max}

\newtheorem{lemma}{Lemma}
\newtheorem{proposition}{Proposition}
\newtheorem{remark}{Remark}

\newtheorem{theorem}{Theorem}

\newcommand{\mytitle}{Stochastically Dominant Distributional Reinforcement Learning}
\icmltitlerunning{\mytitle}

\begin{document}

\twocolumn[
\icmltitle{\mytitle}



\icmlsetsymbol{equal}{*}

\begin{icmlauthorlist}
\icmlauthor{John D. Martin}{stev}
\icmlauthor{Michal Lyskawinski}{stev}
\icmlauthor{Xiaohu Li}{stev}
\icmlauthor{Brendan Englot}{stev}
\end{icmlauthorlist}

\icmlaffiliation{stev}{Stevens Institute of Technology, Hoboken, New Jersey, USA}

\icmlcorrespondingauthor{John D. Martin}{jmarti3@stevens.edu}

\icmlkeywords{reinforcement learning, aleatoric uncertainty, risk-aware control, off-policy learning}

\vskip 0.3in
]



\printAffiliationsAndNotice{}  

\begin{abstract}
	We describe a new approach for managing aleatoric uncertainty in the Reinforcement Learning (\textsc{rl}) paradigm. Instead of selecting actions according to a single statistic, we propose a distributional method based on the second-order stochastic dominance (\textsc{ssd}) relation. This compares the inherent dispersion of random returns induced by actions, producing a comprehensive evaluation of the environment's uncertainty. The necessary conditions for \textsc{ssd} require estimators to predict accurate second moments. To accommodate this, we map the distributional \textsc{rl} problem to a Wasserstein gradient flow, treating the distributional Bellman residual as a potential energy functional. We propose a particle-based algorithm for which we prove optimality and convergence. Our experiments characterize the algorithm's performance and demonstrate how uncertainty and performance are better balanced using an \textsc{ssd} policy than with other risk measures. 	
\end{abstract}

\section{Introduction}
Often in Reinforcement Learning (\textsc{rl}), agents select actions to maximize their expected sum of future (discounted) rewards. Many have pointed out how this strategy will sometimes lead to undesirable outcomes, particularly when the environment is stochastic \cite{riskrl:1994:heger:consideration_of_risk_in_reinforcement_learning, riskrl:2013:tamar_dicastro_mannor:temporal_difference_methods_for_the_variance_of_the_reward_to_go,riskrl:2020:keramati_etal:being_optimistic_rl}. In these domains, an interesting subclass of problems require the agent to decide between several competing alternatives with the same expected outcome. These scenarios frequently arise in finance \cite{riskmod:2006:dentcheva:portfolio_opt}, where mutliple portfolios can lead to the same return but with different variability. In such settings, the expected value does not capture the full state of uncertainty, and it becomes prudent to employ the full distribution of outcomes. 

The Conditional Value at Risk ($\textnormal{CVaR}_{\alpha}$) is a popular statistic that measures uncertainty with the expected outcome under an $\alpha$-fraction of possibilities \cite{riskmod:1999:artzner_etal:cvar}. A great deal of recent \textsc{rl} research focuses on learning good CVaR policies \cite{riskrl:2014:chow_algorithms_for_cvar_optimization_in_mdps, riskrl:2015:tamar_etal:opt_cvar, distrl:2018:dabney_etal:implicit_quantile_networks_for_distributional_reinforcement_learning, riskrl:2020:keramati_etal:being_optimistic_rl}. One point that remains unresolved is how to choose the right fraction of outcomes, i.e. the risk level $\alpha$. It seems plausible that the best $\alpha$-subset could vary across the environment. To our knowledge no one has considered this problem in \textsc{rl}, when uncertainty is used to evaluate competing actions. To address these issues, we introduce a distributional policy that simultaneously captures many risk levels, therefore removing the need to select one. Our policy is based on the Second Order Stochastic Dominance (\textsc{ssd}) relation.

The \textsc{ssd} relation is defined using distribution functions and compared over the continuum of their realizable values. We say that $X$ stochastically dominates $Y$ in the second order when their cumulative CDFs, $F^{(2)}(\alpha)=\int_{-\infty}^\alpha F(x)dx$, satisfy \eqref{eq:ssd}, and we denote the relation $X\succeq_{(2)} Y$:
\begin{align}\label{eq:ssd}
	X\succeq_{(2)} Y\Longleftrightarrow F^{(2)}_X(\alpha) \leq F^{(2)}_Y(\alpha) \ \forall \alpha \in \mathbb{R}.
\end{align}
The function $F^{(2)}$ defines the frontier of what is known as the \textit{dispersion space} (Figure \ref{fig:dispersion}). The volume reflects the degree to which a random variable differs from its expected value, or its deterministic behavior. Outcomes that are disperse have more uncertainty and are considered risky in decision making settings. Indeed, a fundamental result from expected utility theory states that rational risk-averse agents prefer $X$ to $Y$ when $X\succeq_{(2)} Y$ \cite{sdom:2006:dentcheva_ruszcynski:inverse_stochastic_dominance_constraints_and_rank_dependent_expected_utility_theory}. Drawing inspiration from this strategy, we apply \textsc{ssd} as an action selection method to reduce dispersion in the data distribution with which a policy is learned. Our paper offers the following contributions:
\begin{figure}
	\centering	
	\includegraphics[width=0.95\columnwidth]{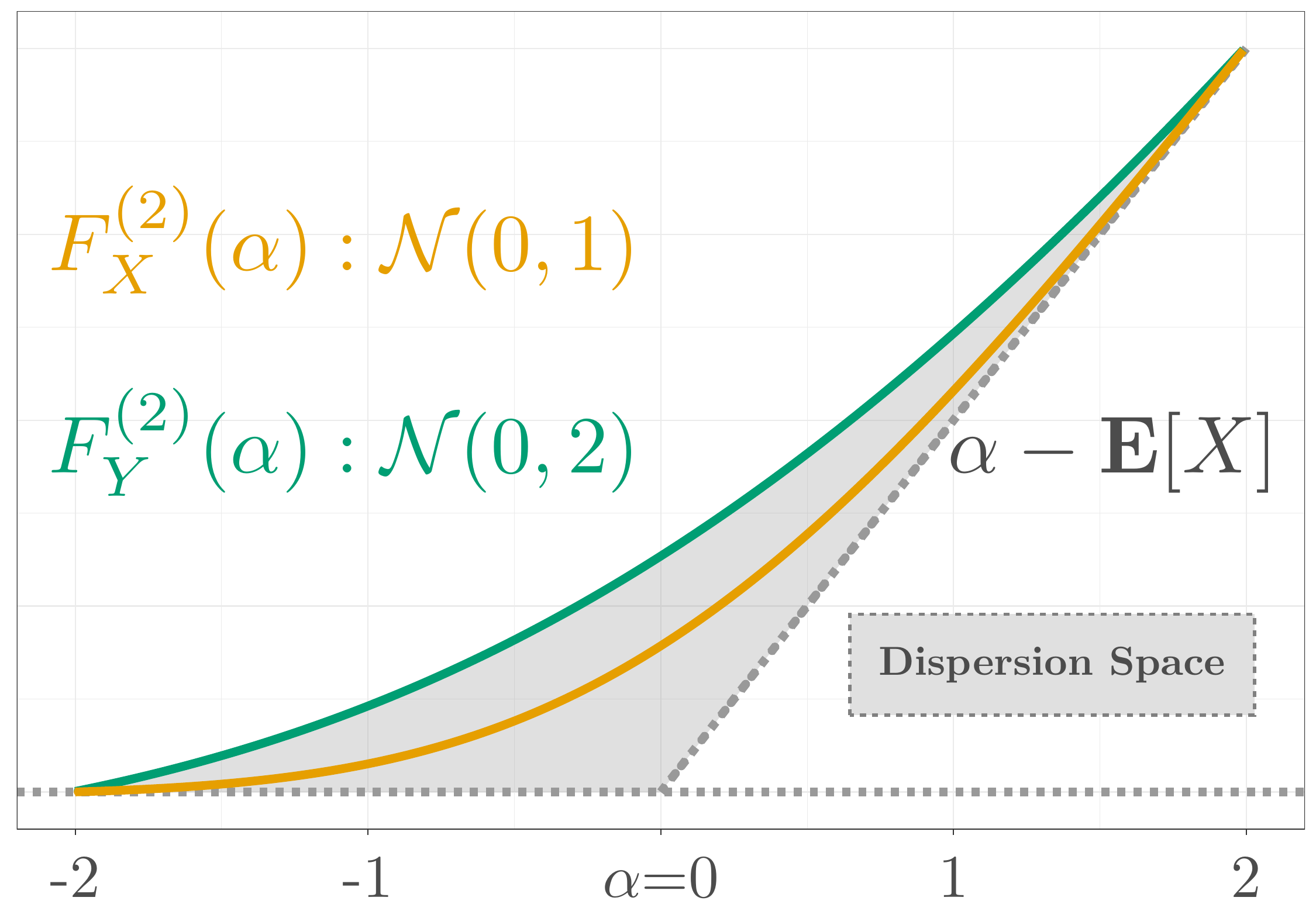}
	\caption{\textbf{Dispersion space:} The relative uncertainty of a random variable is shown as the space between its cumulative CDF $F^{(2)}_X$ and the asymptotes (dotted). Here, the line $\alpha-\Ebf[X]$ defines the behavior of $X$ as its uncertainty vanishes.}
	\label{fig:dispersion}
\end{figure}

\paragraph{A distributional policy:} Metrics such as variance \cite{riskrl:2001:sato_kimura_kobayashi:td_algorithm_for_the_variance_of_return_and_mean_variance_reinforcement_learning,riskrl:2013:tamar_dicastro_mannor:temporal_difference_methods_for_the_variance_of_the_reward_to_go} and quantile statistics, like CVaR, \cite{riskrl:2014:chow_algorithms_for_cvar_optimization_in_mdps, distrl:2018:dabney_etal:implicit_quantile_networks_for_distributional_reinforcement_learning, riskrl:2020:keramati_etal:being_optimistic_rl} are prevalent in \textsc{rl}. A novel contribution of our work is the introduction of \textsc{ssd} for distributional action selection. As we will show, this relation eliminates the need to tune risk parameters. We apply the relation in settings where there are many solutions, and the agent wishes to select the least disperse (i.e. most certain) option.  
 
\paragraph{A new distributional \textsc{rl} algorithm:}\textsc{Ssd} implies an ordering on the first two moments of distributions. We propose a new learning algorithm based on Wasserstein gradient flows that is guaranteed to respect this, because its estimates converge in the first two moments.

We validate our theoretical claims with several targeted experiments. The main hypothesis we test is that the \textsc{ssd} policy induces the least-disperse data distribution from which optimality can be achieved when learning off-policy.

\section{Background}
\begin{figure*}
	\centering
	\begin{subfigure}[b]{0.3\textwidth}
        \includegraphics[width=\textwidth]{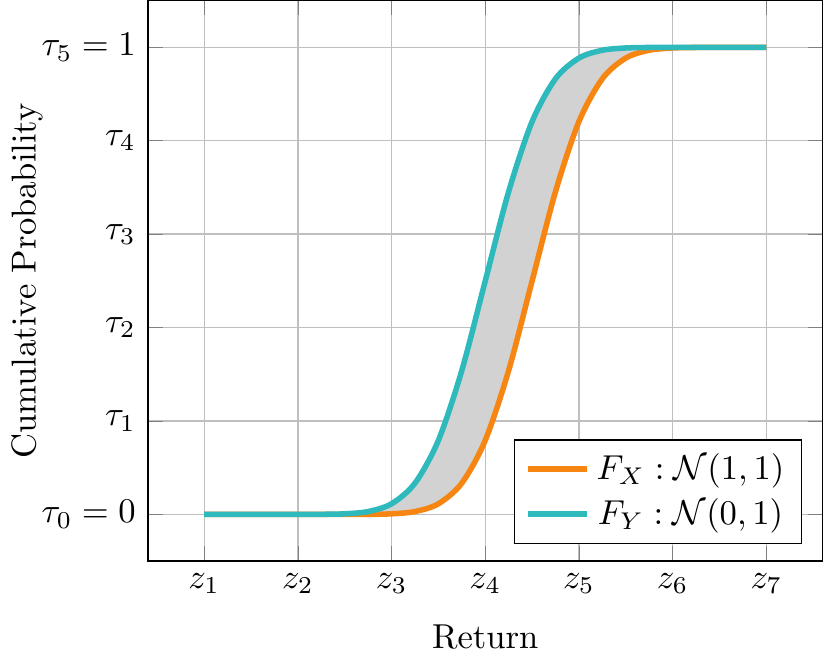}	
        \caption{$X\succeq_{(1)}Y$ and $X\succeq_{(2)}Y$}
        \label{fig:xfsdy}
    \end{subfigure}
    \begin{subfigure}[b]{0.3\textwidth}
        \includegraphics[width=\textwidth]{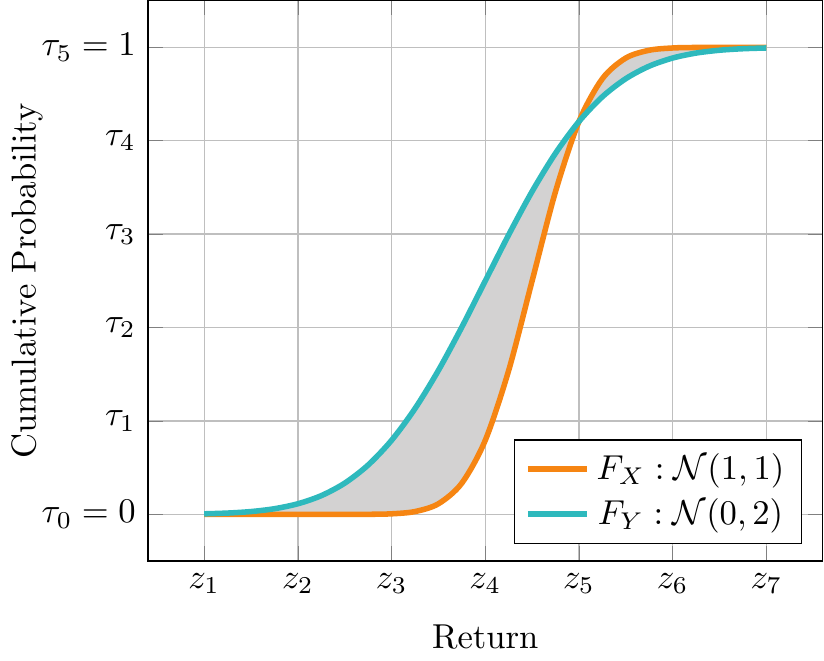}	
        \caption{$X\nsucceq_{(1)}Y$ and $X\succeq_{(2)}Y$}
        \label{fig:xssdy}
    \end{subfigure}
    \begin{subfigure}[b]{0.3\textwidth}
        \includegraphics[width=\textwidth]{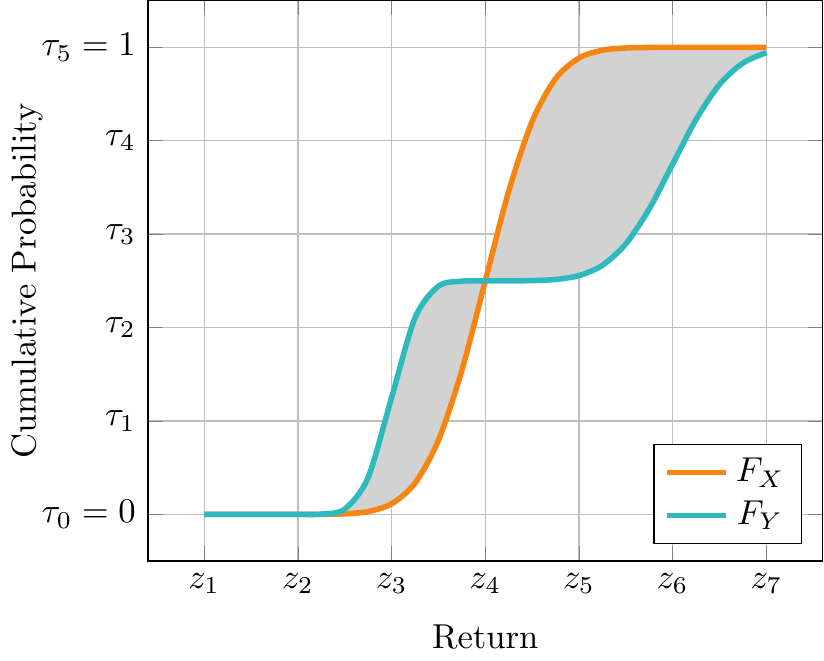}	
        \caption{$X\nsucceq_{(1)}Y$ and $X\nsucceq_{(2)}X$}
        \label{fig:no_ssd}
    \end{subfigure}
	\caption{\textbf{Stochastic dominance action selection:} Imagine $X$ and $Y$ are returns induced by different actions. Our action selection rule can be visualized with plots of the CDF. In Fig. \ref{fig:xfsdy} $X\succeq_{(2)}Y$, because $X$ places more mass on points larger than $\alpha$. In Fig. \ref{fig:xssdy}, the area left of $z_5$ is greater than the area to its right, hence $X\succeq_{(2)}Y$. In Fig. \ref{fig:no_ssd}, neither variable dominates, because for $\alpha \geq z_4$, the enclosed area is larger than the other region. In these cases, we select from among the competing actions at random.}
	\label{fig:ssd_policy}
\end{figure*}
Reinforcement Learning describes a sequential decision making problem, whereby an agent learns to act optimally from rewards collected after taking actions.  At each time step, the agent selects an action $A\in\Acal$ based on its current state $S\in\Scal$, then transitions to the next state $S'\in\Scal$ and collects a reward $R\in\mathbb{R}$. The interaction is formally modeled as a Markov Decision Process (MDP) \cite{classic:1994:putterman:mdp}, which we denote $\left<\Scal,\Acal,p, \gamma\right>$. The transition kernel $p\colon \Scal\times\Acal \rightarrow \Pscr(\mathbb{R}\times \Scal)$ defines a joint distribution over the reward and next state, given the current state-action pair. Here, $\Scal$ and $\Acal$ are measurable Borel subsets of complete and separable metric spaces, which we take as finite. And for each state $s\in\Scal$, the set $\Acal_s \subset \Acal$ is a measurable set indicating the feasible actions from $s$. The \textit{random return} is
 \begin{align}\label{eq:return}
 	Z_\pi^{(s,a)} &\triangleq \sum_{t=0}^\infty \gamma^t R^{(S_t,A_t)} \ \biggl| \ S_0 = s,A_0 = a.
 \end{align}
 Returns reflect the outcome of a decision sequence that starts after taking action $a$ in state $s$ then following the policy $\pi\in \Pi$ thereafter. Policies are stationary distributions over actions, coming from the set $\Pi=\{\pi | \pi \colon \Scal \rightarrow \Pscr(\Acal) \}$. Here, $\gamma \in [0,1)$ is a discount factor, and $R^{(S_t,A_t)}$ is the real-valued random reward associated with the state and action at time $t\in\mathbb{N}$. 
 
\subsection{Bellman's Equations}
\citet{dp:1957:bellman:dynamic_programming} showed that the expected return \eqref{eq:return}, also known as the \textit{value function}, has a recursive decomposition:
\begin{align*}
	 Q_\pi^{(s,a)} &\triangleq \Ebf_{\mu^{(s,a)}_\pi}[Z_\pi^{(s,a)}]= \Ebf_{p,\pi}[R +  \gamma Q_\pi^{(S',A')}],\\
	 Q_*^{(s,a)} &\triangleq \Ebf_{\mu^{(s,a)}}[Z^{(s,a)}]= \Ebf_{p}[R +  \gamma \max_{a'\in\Acal_{S'}}Q_*^{(S',a')}].
\end{align*}
Here the \textit{value}, $Q_\pi^{(s,a)}$, is defined as the expected return \cite{rl:1998:sutton_barto:introduction_to_reinforcement_learning} of policy $\pi$. We denote $\mu^{(s,a)}_\pi$ to be the corresponding distribution of returns under $\pi$ from $(s,a)$. Returns under $\mu^{(s,a)}$ follow the greedy policy, $\pi_*(s) = \argmax_{a\in \Acal_s} \Ebf_{\mu^{(s,a)}}[Z^{(s,a)}]$ for all $s\in\Scal$. When clear from context, we will drop the superscripts and refer to a single measure $\mu^{(s,a)}$ as $\mu$. Viewed as a functional operator, these equations are known to contract to a unique fixed point. The contractive property motivates algorithms that update representations of $Q_*$ to minimize the difference formed between both sides. Two popular methods for model-free learning are Sarsa \cite{rl:1994:rummery_niranjan:online_q_learning_using_connectionist_systems} and $Q$-learning \cite{rl:1992:watkins:qlearning}. These algorithms use samples of the form $(s,a,r,s')$ to iteratively update value estimates. Sarsa is an on-policy algorithm because it evaluates the policy it uses to gather data:
\begin{align*}
	Q_\pi^{(s,a)} \gets Q_\pi^{(s,a)} + \alpha(r-\gamma \Ebf_{\pi}[Q_\pi^{(s',A')}]- Q_\pi^{(s,a)}).
\end{align*}
$Q$-learning is \textit{off-policy} because it gathers data with a separate \textit{behavior policy} to learn the target greedy policy:
\begin{align*}
	Q_*^{(s,a)} \gets Q_*^{(s,a)} + \alpha(r-\gamma \max_{a'\in\Acal_{s'}} Q_*^{(s',A')}- Q_*^{(s,a)}).
\end{align*}
\subsection{Distributional Bellman Operators}
The return distribution, $\mu^{(s,a)}_\pi$, also satisfies a distributional variant of Bellman's equation $\Tcal^\pi\mu^{(s,a)}_\pi \triangleq$
 \begin{align}\label{eq:distop}
 	\int_{\mathbb{R}}\sum_{(s',a')\in\Scal\times\Acal}f^{(r,\gamma)}_{\sharp}\mu^{(s',a')}_\pi\pi(a'|s')p(dr,s'|s,a).
 \end{align}
Here, $\Tcal^\pi$ is the distributional Bellman operator. It embeds a measurable mapping that reflects the Bellman action: $f^{(r,\gamma)}(x) \triangleq r+\gamma x$, where the push-forward measure is $f_\#(\mu(A)) \triangleq \mu(f_\#^{-1}(A))=\nu(A)$, for all Borel measurable sets $A$. Just as the standard Bellman equation is the focus of standard value-based RL, the distributional Bellman operator \eqref{eq:distop} plays the central role in Distributional RL (\textsc{drl}); it motivates algorithms which attempt to represent $\mu^{(s,a)}_\pi$ and approximate it by repeated application of the update $\mu_{\pi,t+1}^{(s,a)} \gets \Tcal^\pi\mu_{\pi,t}^{(s,a)}$, for any $(s,a)\in\Scal\times\Acal$. The optimality operator is realized under the greedy policy:
 \begin{align}\label{eq:control_op}
 	\Tcal\mu^{(s,a)} &\triangleq\int_{\mathbb{R}}\sum_{(s',a')\in\Scal\times\Acal}f^{(r,\gamma)}_{\sharp}\mu^{(s',a^*)}p(dr,s'|s,a).
 \end{align}
 
\section{Distributional Action Selection}\label{sec:rl_free_energy}
We consider scenarios where a \textsc{drl} agent often encounters several outcomes which all appear equivalent under the expected return. The uncertainty-sensitive decision problem is to select from among these choices, the option that minimizes uncertainty. For this we propose the \textsc{ssd} policy, whose comparisons are visualized in Figure \ref{fig:ssd_policy}.

At each state $s$, the agent makes a point-wise comparison of the distribution functions $F^{(2)}_{Z^{(s,a)}}(\alpha)$, for all $a\in\Acal_s$. The dominating action is the one whose cumulative CDF achieves the lowest value for every $\alpha\in\mathbb{R}$:
\begin{align*}
\Acal^{(2)}_s\triangleq\{ a \in \Acal_s : Z^{(s,a)} \succeq_{(2)} Z^{(s,a')}, \forall \ a' \in \Acal_s \setminus \{a\}\}.
\end{align*}
\subsection{Numerically tractable comparisions}
Evaluating \textsc{ssd} appears to be a numerically intractable task, involving point-wise comparisons over an infinite domain. Fortunately, we can circumvent this problem by using cumulative quantile functions \cite{sdom:2006:dentcheva_ruszcynski:inverse_stochastic_dominance_constraints_and_rank_dependent_expected_utility_theory}: $F^{-2}(\tau)=\int_0^\tau F^{-1}(t)dt$. Now the \textsc{ssd} relation becomes: 
\begin{align}\label{eq:invsd}
X\succeq_{(2)}Y \Longleftrightarrow F^{-2}_X(\tau) \geq F^{-2}_Y(\tau) \ \forall \ \tau \in (0,1),	
\end{align}
where we assume that $F^{-2}_Y(0)=0$, and $F^{-2}_Y(1)=\infty$. 

Notice that $F^{-2}_X(\tau)/\tau = \Ebf[X|X\leq \xi^{(\tau)}]$ is the Conditional Value at Risk for level $\tau$.  Thus, the \textsc{ssd} relation can be interpreted as a continuum of CVaR comparisons for all $\tau\in(0,1)$. 
 From this we can surmise that points along the boundary of dispersion space (Figure \ref{fig:dispersion}) represent unconditional Values at Risk (VaR). Furthermore, this connection suggests a numerically-tractable way to compute $F^{-2}$. 
\begin{lemma}\label{lem:ssd}
	Let $\tau \in (0,1)$ and consider $\xi^{(\tau)} = F^{-1}_X(\tau)$. Then $F^{-2}_X(\tau) =\Ebf[X\leq \xi^{(\tau)}]$.
\end{lemma}
Lemma \ref{lem:ssd} makes it possible to compare total expectations on subsets of the return space, instead of dealing with probability integrals over an unbounded domain.

Computations simplify even further when we consider discrete measure approximations to the return distribution. We propose a Lagrangian (particle-based) discretization, where the measure is supported on $N\in\mathbb{N}$ equally-likely diracs: 
\begin{align*}
	\mu^{(s,a)} \approx \frac{1}{N}\sum_{i=1}^N\delta_{z^{(i)}}^{(s,a)}. 
\end{align*}
Values are straightforward to compute from the corresponding samples: $Q^{(s,a)} = \frac{1}{N}\sum_{i=1}^Nz^{(i)}$. 

To apply \eqref{eq:invsd}, denote the ordered coordinates of a return distribution to be $z^{[1]} \leq z^{[2]}\leq \cdots \leq z^{[N]}$. Then given particle sets for two random returns induced by the actions $a_1$ and $a_2$, we have the following result.
\begin{proposition} 
$Z^{(s,a_1)} \succeq_{(2)} Z^{(s,a_2)}$ if, and only if
\begin{align}
	\sum_{i=1}^jz^{[i]}_{a_1} &\geq \sum_{i=1}^jz^{[i]}_{a_2}, \ \forall \ j=1,\cdots,N.\label{eq:finite_ssd}
\end{align}
\end{proposition}

The \textsc{ssd} policy is executed at each step by constructing $\Acal^{(2)}_s$ using \eqref{eq:finite_ssd} and a discrete representation of $\mu^{(s,a)}$. In some cases $\Acal^{(2)}_s$ will be empty (Figure \ref{fig:no_ssd}), indicating that total dominance cannot be established. There are several heuristics that could handle this outcome, including a next-best strategy, or an additional decision criterion. We choose to sample the greedy actions uniformly at random. This increases uniform exploration when dominance cannot be established and still constitutes a strict enhancement of the greedy policy when multiple solutions are present. 

\subsection{Necessary conditions for \textsc{ssd}}
When is it possible to apply the \textsc{ssd} policy? The following result from \citet{sdom:1980:fishburn:stochastic_dominance_and_moments_of_distributions} implies an ordering on the first two moments of the distributions under \textsc{ssd}.
\begin{proposition}[\citet{sdom:1980:fishburn:stochastic_dominance_and_moments_of_distributions}]\label{prop:moment_ordering}
	Assume $\mu$ has two finite moments. Then $X\succeq_{(2)} Y$ implies $\mu_X^{(1)} \geq \mu_Y^{(1)} $ or $\mu_X^{(1)} = \mu_Y^{(1)}$ and $\mu_X^{(2)} \leq \mu_Y^{(2)} $, where $(\cdot)$ denotes a particular moment of the distribution $\mu$.   
\end{proposition}
The ordering indicates that the dominating distribution either has the greatest mean, or it has the smallest second moment when means are equal. Given infinite precision and random initialization, the chances of more than one action having the same value may seem unlikely. However, in cases where finite precision is used (e.g. finance), or cases where a tolerance is applied to comparisons, equivalence arises often. Proposition \ref{prop:moment_ordering} imposes a necessary requirement on estimates of the return distribution. Namely, the estimates must be accurate enough to respect the ordering. Moving forward we seek distributional learning algorithms that we know converge in the first two moments. 

\section{Wasserstein Gradient Flows for RL}
In this section we describe how return distributions can be obtained from the solution of a Wasserstein Gradient Flow (\textsc{wgf}). We detail the solution procedure and show how it naturally integrates into the fitted value iteration paradigm \cite{rl:1995:gordan:fittedvi}. We expand on the \textsc{wgf} theory to show that solutions converge in the first two moments, as we need to respect Proposition \ref{prop:moment_ordering}. 
  
\subsection{Wasserstein convergence}
The $k$-th order Wasserstein distance for any two univariate measures $\mu,\nu\in\Pscr_k(\mathbb{R})$,  is defined as
\begin{align*}
	\Wscr_k(\mu,\nu) &\triangleq \inf_{\gamma\in\Pscr_k(\mu,\nu)}\left\{\int_{\mathbb{R}^2}|x-y|^kd\gamma(x,y) \right\}^{1/k},
\end{align*}
where $\Pscr_k(\mu,\nu)$ is the set of all joint distributions with marginals $\mu$ and $\nu$ having $k$ finite moments. The distance describes an optimal transport problem, where one seeks to transform $\mu$ to $\nu$ with minimum cost; here the cost is $|x-y|^k$. The $\Wscr_k$ distance is appealing as a distributional learning objective, because its convergence implies convergence in the first $k$ moments \cite{otrans:2008:villani:optimal_transport_old_and_new}.

\subsection{Distributional RL as free-energy minimization}
We cast the distributional \textsc{rl} problem as a free-energy minimization in terms of the functional: 
\begin{align}\label{eq:free_energy_rl}
	E(\mu)&\triangleq F(\mu) +\beta^{-1}H(\mu).
\end{align}
Here, we have dropped the superscript notation. $F$ is the potential and $H$ is the entropy of a single probability measure; $\beta \in \mathbb{R}_+$ is an inverse temperature parameter. 

The potential energy defines what it means for a distribution to be optimal. We choose the low-energy equilibrium to coincide with minimum expected Bellman error, formed from \eqref{eq:control_op}. Energy is minimized when the mapping $\Tcal$ reaches its fixed point, $\Tcal\mu^{(s,a)} = \mu^{(s,a)}$ for some $(s,a)$. Given a transition sample $(s,a,r,s')$, we compute the distributional targets $\Tcal z^{(s,a)}$, which denote realizations of $\Tcal\mu^{(s,a)}$, and define Bellman's potential energy as
 \begin{align}\label{eq:bellman_potential}
	F(\mu) &\triangleq \frac{1}{2}\int\left(\Tcal z^{(s,a)} - z^{(s,a)}\right)^2d\mu=\int U(z)d\mu.
\end{align}
The optimal probability measure for these models is known to be the Gibbs measure: $\mu_*(z) = \Zcal^{-1}\exp\{-\beta U(z)\}$, where $\Zcal= \int\exp\{-\beta U(z)\}dz$. Energy-based models have been applied for policy optimization \cite{energyrl:2017:haarnoja_etal:reinforcement_learning_with_deep_energy_based_policies, gflow:2018:zhang_etal:policy_optimization_as_wasserstein_gradient_flows}, but to our knowledge, they have not appeared in value-based methods for \textsc{drl}.

\subsection{The Fokker-Planck Equation}
We would like to understand the convergence behavior of return distributions as the free-energy is minimized. Systems of this nature are typically modeled as continuous-time stochastic diffusion processes, where the distributions $\{\mu_t\}_{t\in[0,1]}$ evolve over a smooth manifold of probability measures from $\Pscr_2(\mathbb{R})$. The dynamics of $\mu_t$ is known to obey a diffusive partial differential equation called the Fokker-Planck equation \cite{fp:1984:risken:fp}:
\begin{align}\label{eq:fp}
	\partial_t\mu_t  = \nabla\cdot\left(\mu_t \nabla(\frac{\delta E}{\delta \mu_t})\right).
\end{align}
Here, the sub-gradient with respect to time is denoted $\partial_t$, and the first variation (G\^ateaux derivative) of free energy $\frac{\delta E}{\delta \mu}$. The Fokker-Plank equation plays a central role in statistical physics, chemistry, and biology. In optimization, it defines the solution path, or gradient flow, of $\mu$ as it evolves over the manifold of probability measures.
\begin{proposition}[\citet{gflow:2005:ambrosio:gradient_flows_in_metric_spaces_and_in_the_space_of_probability_measures}]
	Let $\{\mu_t\}_{t\in[0,1]}$ be an absolutely-continuous curve in $\Pscr_2(\mathbb{R})$. Then for $t\in[0,1]$, the vector field $\vbf_t = \nabla(\frac{\delta E}{\delta t}(\mu))$ defines a gradient flow on $\Pscr_2(\mathbb{R})$ as $\partial_t \mu_t =- \nabla \cdot(\mu_t\vbf_t)$, where $\nabla\cdot \ubf$ is the divergence of the vector $\ubf$.
\end{proposition}
Intuitively, the free-energy $E$ characterizes the diffusion process and thus, the optimization landscape of our new distributional RL problem. Convergence to an optimal point can be guaranteed provided $E$ is convex, which we know to be the case for \eqref{eq:free_energy_rl}, which is quadratic and logarithmic in $\mu$.  

\subsection{Discrete Time Solutions}
To approximately solve \eqref{eq:fp}, we adopt an iterative procedure due to \citet{gflow:1998:jordan_etal:the_variational_formulation_of_the_fokker_planck_equation}. The method discretizes time in steps of $h\in\mathbb{R}_+$ and applies the proximal operator
\begin{align}\label{eq:jko}
	\text{Prox}_{hE}^{\Wscr}(\mu_k) \triangleq \argmin_{\mu\in\Pscr_2(\mu,\mu_k)} \Wscr^2_2(\mu,\mu_k)+2h E(\mu).
\end{align}
For every step $k\in\mathbb{N}$, the operator generates a path of distributions $\{\mu_t\}_{t=1}^K$ such that  $\mu_{k+1} = \text{Prox}_{hE}^{\Wscr}(\mu_k)$ is equivalent to $\mu_{K}$. In contrast with \textsc{drl} methods that apply \eqref{eq:control_op}, we apply the proximal operator to minimize a free energy with a $\Wscr_2^2$-regularizer via (semi-)gradient steps. And because $E$ is convex, this method obtains the unique solution to \eqref{eq:fp}.
\begin{proposition}[\citet{gflow:1998:jordan_etal:the_variational_formulation_of_the_fokker_planck_equation}]\label{lemma:ito_conv}
	Let $\mu_0\in\Pscr_2(\mathbb{R})$ have finite free energy $E(\mu_0)<\infty$, and for a given $h>0$, let $\{\mu^{(h)}_t\}_{t=1}^K$ be the solution of the discrete-time variational problem \eqref{eq:jko}, with measures restricted to $\Pscr_2(\mathbb{R})$, the space with finite second moments. Then as $h\rightarrow 0$, $\mu_K^{(h)} \rightarrow \mu_T$, where $\mu_T$ is the unique solution of \eqref{eq:fp} at $T=hK$.
\end{proposition}
Furthermore, one can evaluate the free-energy \eqref{eq:bellman_potential} over the solution sequence and observe it becomes a decreasing function of time (i.e. a Lyapunov function). The result implies that the expected distributional Bellman residual is minimized when using the JKO approach. 
\begin{proposition}\label{prop:lyapunov_energy}
	Let $\{\mu^{(h)}_k\}_{k=0}^K$ be the solution of the discrete-time variational problem \eqref{eq:jko}, with measures restricted to $\Pscr_2(\mathbb{R})$, the space with finite second moments. Then $E(\mu_k)$ is a decreasing function of time. 
\end{proposition}

Finally, we can show that as $\beta$ is annealed, the output of our free-energy optimization \eqref{eq:jko} is equivalent to the solution obtained from the distributional Bellman operator \eqref{eq:control_op}. 
\begin{theorem}
	If $\Tcal\mu=\mu$, then $\textnormal{Prox}_{hE}^{\Wscr}(\mu)=\mu$ as $\beta \rightarrow \infty$.
\end{theorem}


\subsection{Discrete Measure Solutions} \label{sec:discrete_measure}
Given an initial set of particles at some state-action pair $z(s,a) = \{z^{(1)},\cdots,z^{(N)}\}$, we evolve them forward in time with steps of $h$ to obtain the solution at $t+h$. We apply a finite number of gradient steps to approximate the convergence limit $T=hK$. Finally we consider an entropic-regulated form of $\Wscr_2^2$ \citep{otrans:2013:cuturi:sinkhorn_distances_lightspeed_computation_of_optimal_transport} for two finite distributions $\mu = \sum_{i=1}^N\mu_i\delta_{x^{(i)}}$ and $\nu = \sum_{j=1}^N\nu_j\delta_{y^{(j)}}$: 
\begin{align*}
	\Wscr_{\beta}(\mu,\nu) &\triangleq \inf_{P\in \mathbb{R}^{N\times N}_+} \left<P,C\right> + \beta \mathsf{KL}(P|\mu\otimes \nu), \\
	&\text{s.t.} \ \sum_{j=1}^N P_{ij} = \mu_i, \sum_{i=1}^N P_{ij} = \nu_j.
\end{align*}
Here, $\left<P,C\right>$ denotes the Frobenius norm between the joint $P$ and the square Euclidean cost $C_{ij} = (x_i-y_j)^2$, and $\mathsf{KL}(P|\mu\otimes \nu) = \sum_{i,j}[P_{ij}\log(P_{ij}/\mu_{i}\nu_j)-P_{ij}+\mu_i\nu_j]$. The entropic term promotes numerical stability by acting as a barrier function in the positive octant. JKO stepping under this new distance is denoted
\begin{align}\label{eq:prox_2}
	\text{Prox}_{hF}^{\Wscr_\beta}(\mu_k)&\triangleq \argmin_{\mu\in \Pscr_2(\mu,\mu_k)}\Wscr_{\beta}(\mu,\mu_k) + 2h F(\mu).
\end{align}
One can compute the entropic-regularized distance, $\Wscr_\beta$, using Sinkhorn iterations \citep{otrans:1967:sinkhorn:diagonal_equivalence_to_matrices_with_prescribed_row_and_column_sums}. This procedure (detailed in the appendix) is differentiable, which allows us to update represent particle locations with parametric models and update their predictions with gradient steps computed through auto-differentiation.

\subsection{Online WGF Fitted $Q$-iteration}
We are now ready to describe Online WGF Fitted $Q$-iteration (Alg. \ref{alg:dpa}). The algorithm combines the solution of \eqref{eq:prox_2} into a Fitted $Q$-iteration framework to repeatedly fit return distributions. The loss is computed with Alg. \ref{alg:proximal_step}. The principles apply in both the on-policy and off-policy settings. Here, we consider the off-policy case to compare different behavior policies. Both distributional policies and those based on point estimates are represented with the operator $\Bcal \colon \Pscr(\mathbb{R})^{|\Acal|}\rightarrow\Pscr(\Acal)$. Given a set of return distributions, this outputs a distribution over actions.
\begin{algorithm}[H]
	\caption{Online WGF Fitted $Q$-iteration}
	\label{alg:dpa}
	\begin{algorithmic}[1]
	\STATE {\color{gray} \# Initialize particles}
	\STATE $z(s,a) = \{z^{(i)}\}^N_{i=1}$ $\forall$ $(s,a)\in \Scal\times \Acal$
	\FOR{$t=1,2,\cdots$}
		\STATE {\color{gray} \# Explore with the behavior policy}
		\STATE $s', r \sim p(\cdot | s, a)$ with $a \sim \Bcal z(s,:)$
		\STATE {\color{gray} \# Exploit with the greedy target policy}
		\STATE $a^* \gets \argmax_{a\in\Acal}\{ \frac{1}{N}\sum_{i=1}^N z^{(i)}(s',a)\}$
		\STATE $\Tcal z^{[i]}  \gets r + \gamma z^{[i]}(s',a^*)$ $\forall$ $i\in [N]$
		\STATE {\color{gray} \# Update particles with proximal step}
		\STATE $z(s,a)\gets \argmin_{z}L^{\Wscr_\beta}_{hF_{\Tcal}}(z,z(s,a))$
	\ENDFOR
	\end{algorithmic}
\end{algorithm}
\vspace{-1em}
\begin{algorithm}[H]
	\caption{Proximal Loss}
	\label{alg:proximal_step}
	\begin{algorithmic}[1]
	\STATE \textbf{input:} Source and target particles $z, z_0;\Tcal z$
	\STATE $F_{\Tcal}(z) \gets \frac{1}{2N}\sum_{i=1}^N[\Tcal z^{[i]}-z^{[i]} ]^2$
	\STATE $\Wscr_\beta(z, z_0) \gets$ Sinkhorn$_\beta(z, z_0)$
	\STATE {\color{gray} \# Output JKO loss}
	\STATE \textbf{output:} $L_{hF_{\Tcal}}^{\Wscr_\beta} = \Wscr_\beta(z, z_0) + 2hF_{\Tcal}(z)$ 
	\end{algorithmic}
\end{algorithm}

\section{Connections with Related Work}
\begin{figure*}
	\centering
	\includegraphics[width=\textwidth]{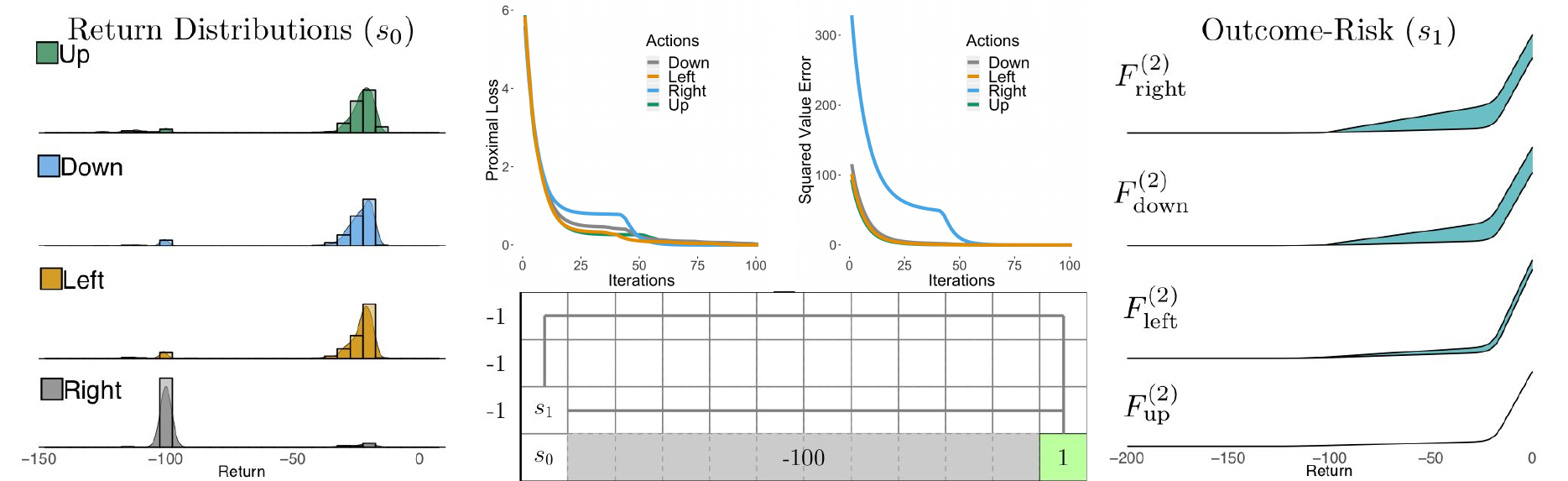}
	\caption{\textbf{Policy evaluation in the CliffWalk domain:} The left plot shows \textsc{wgf} estimates of the smoothed target distributions. Convergence of the proximal loss and the squared value error are shown in the top two plots. Outcome-risk diagrams (right) derived from distribution estimates illustrate  the relative dispersion space size at $s_1$. The two inflections represent the bimodality of the distribution.}\label{fig:cliffwalk}
\end{figure*}

\paragraph{Modeling Risk for \textsc{rl}:}Many have employed measures of uncertainty to replace or regulate the optimization objective in \textsc{rl} using the Markowitz mean-variance model \citep{classic:1952:markowitz:portfolio_selection}. Among these include policy gradient methods \cite{riskrl:2015:tamar_etal:pg_for_corherent_risk_measures}, actor-critics \cite{riskrl:2013:tamar_mannor:var_ac_rl} and TD methods \cite{riskrl:2001:sato_kimura_kobayashi:td_algorithm_for_the_variance_of_return_and_mean_variance_reinforcement_learning, riskrl:2013:tamar_dicastro_mannor:temporal_difference_methods_for_the_variance_of_the_reward_to_go,riskrl:2020:keramati_etal:being_optimistic_rl}. Constraint techniques have also been considered using CVaR within a policy gradient and actor-critic framework \cite{riskrl:2017:chow_et_al:risk_constrained_reinforcement_learning_with_percentile_risk_criteria}. In contrast to methods that directly constrain the policy parameters, we constrain the data distribution with action selection using \textsc{ssd} among the return distributions. \citet{distrl:2018:dabney_etal:implicit_quantile_networks_for_distributional_reinforcement_learning} trains risk-averse and risk-seeking agents from return distributions sampled from various distortion risk measures. However, they do not address problems involving multiple solutions. Furthermore, is it unclear how to sample from \textsc{ssd}-equivalent distortions when total dominance cannot be established. This investigation is left for future work. 

\paragraph{Distributional \textsc{rl}:}Our learning algorithm is inspired by the class of \textsc{drl} algorithms \cite{distrl:2017:bellmare_dabney_munos:a_distributional_perspective_on_reinforcement_learning}. These methods model a distribution over the return, whose mean is the familiar value function, and use it to evaluate and optimize a policy \cite{distrl:2018:barthmaron_etal:distributed_distributional_policy_gradients,distrl:2018:hessel_etal:rainbow_combining_improvements_in_deep_reinforcement_learning}. 
\citet{distrl:2017:bellmare_dabney_munos:a_distributional_perspective_on_reinforcement_learning} first showed the distributional Bellman operator contracts in the supremal Wasserstein distance. They proposed a discrete-measure approximation algorithm (\textsc{c51}) using a fixed mesh in return space and later showed it converges in the Cramer distance \cite{distrl:2018:rowland_bellemare_dabney_monos_teh:an_analysis_of_categorical_distributional_reinforcement_learning}. Particle-based methods that use Quantile Regression (\textsc{qr}), have shown encouraging progress on empirical benchmarks \cite{distrl:2017:dabney_etal:distributional_reinforcement_learning_with_quantile_regression,distrl:2018:dabney_etal:implicit_quantile_networks_for_distributional_reinforcement_learning}. However, understanding their convergence beyond the first moment has been more challenging. By casting the optimization problem as free-energy minimization in the space of probability measures, we show that \textsc{drl} can be modeled as the evolution of a \textsc{wgf}. Updates in this framework have well-defined dynamics, permitting us to better understand convergence and optimality. 

\paragraph{Wasserstein Gradient Flows in RL:}To our knowledge \textsc{wgf} solutions have only been applied to policy gradient algorithms. \citet{gflow:2018:zhang_etal:policy_optimization_as_wasserstein_gradient_flows} models stochastic policy inference as free-energy minimization, and applies the \textsc{jko} scheme to derive a policy gradient algorithm. Their method is couched within the Soft-$Q$ learning paradigm \citep{energyrl:2017:haarnoja_etal:reinforcement_learning_with_deep_energy_based_policies, energyrl:2018:haarnoja_etal:soft_actorcritic_off_policy_maximum_entropy_deep_reinforcement_learning_with_a_stochastic_actor}. These algorithms train a deep neural network to sample from a target Gibbs density using Stein Variational Gradient Descent \citep{vi:2016:liu_wang:stein_variational_gradient_descent_a_general_purpose_bayesian_inference_algorithm}. Our algorithm learns distributions of the underlying return and thus can be considered value-based. Furthermore, we are concerned with decision making in the presence of aleatoric uncertainty, and when the agent must select the most certain outcome from among many alternatives. 

\section{Experiments}
 \begin{figure}
	\centering
	\includegraphics[width=\columnwidth]{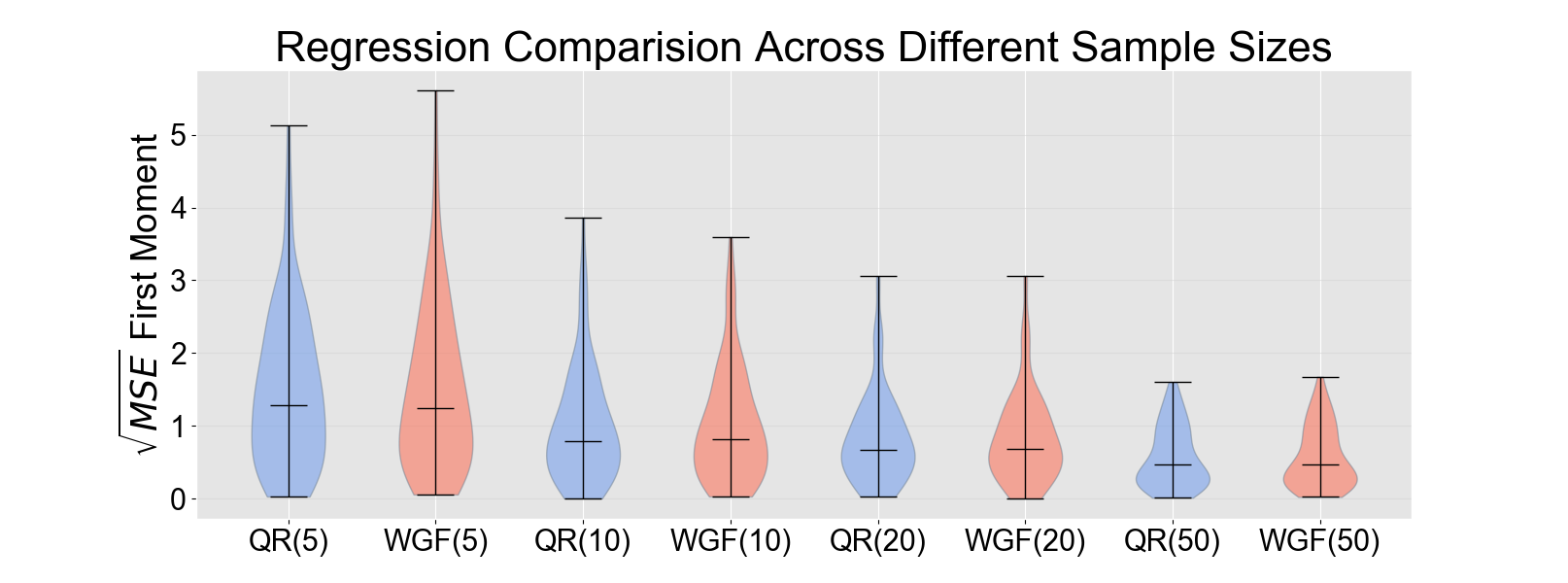}
	\includegraphics[width=\columnwidth]{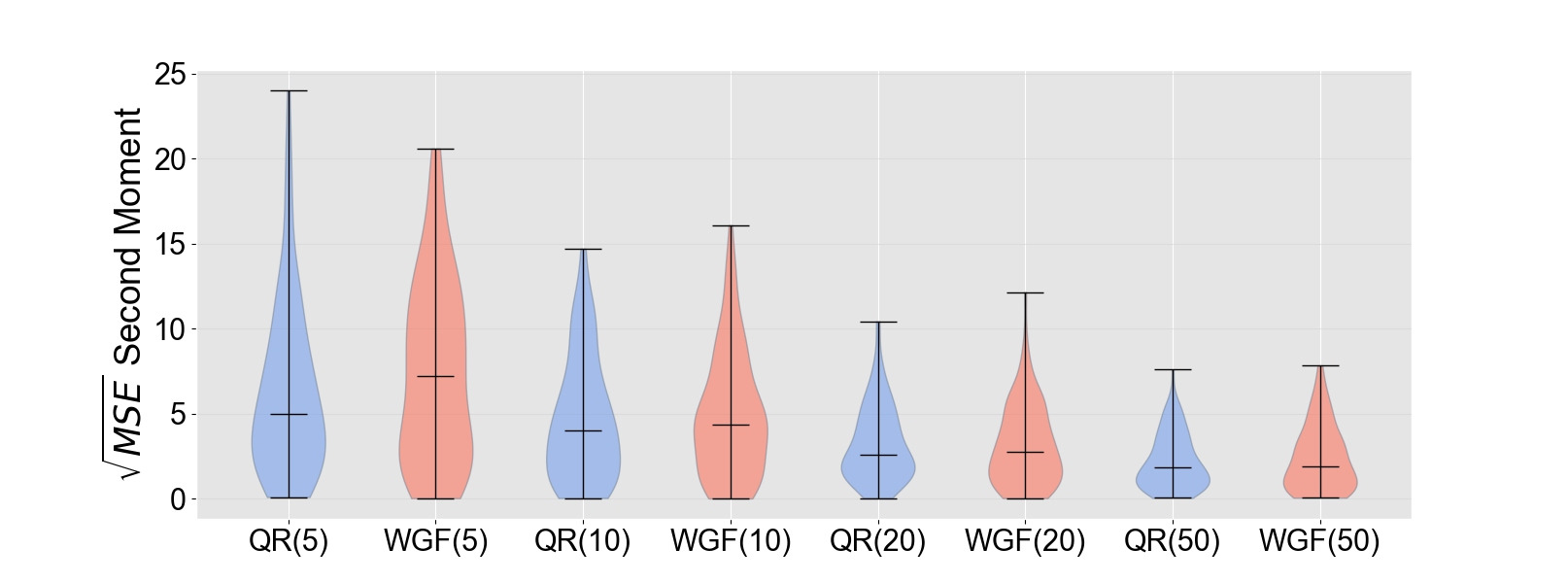}
	\caption{\textbf{Distributions of moment estimation error:} Quantile regression and \textsc{wgf} regression produce similar estimates in one dimension. The number of support samples is shown in parentheses.}
	\label{fig:regression}
\end{figure}

In this section we verify several prior assertions. Namely, we test the hypothesis that \textsc{wgf} regression produces two accurate moment estimates. Next we show \textsc{wgf} solutions from Alg. \ref{alg:dpa} can recover the latent return distribution in a policy evaluation setting. We extend these results to the control setting with bootstrapped off-policy updates under function approximation. In our final experiment, we quantify an agent's ability to mitigate uncertainty while gathering training data with the SSD behavior policy. Details of each experiment can be found in the Appendix. 

\subsection{Regression Comparison}
 Given that standard quantile regression learns samples from a uniform mesh in probability space, theory suggests accuracy improvements can be gained with a non-uniform mesh produced from the solution of a \textsc{wgf}. To evaluate this hypothesis, we compared the root mean squared error on a five component one-dimensional Gaussian mixture model, intended to be representative of a geometrically-complex return distribution. Ablations informed the parameterization of the proximal loss (See appendix). We collected data over 100 independent trials, varying the number of samples each method regressed. Our data shows there to be \textit{no statistical difference} between \textsc{qr} and \textsc{wgf} regression (Figure \ref{fig:regression}). 
 
 We interpret the observed insignificance as a consequence of using low-dimensional data. The error from a uniform grid is expected to become more pronounced as dimensionality increases. Given that we are concerned with one-dimensional return distributions, however, these results inform different message within our problem setting. Namely, the distributions regressed through \textsc{qr} may be reasonably employed for \textsc{ssd} action selection. We believe practitioners will find this result valuable when choosing a regression method where two accurate moment estimates are required.

\subsection{\textsc{wgf} Policy Evaluation}
Proposition \ref{prop:lyapunov_energy} argues that repeated application of the proximal step \eqref{eq:jko} produces a decreasing function of time, implying that the Bellman free energy is minimized at convergence. Here, we verify this is indeed the case by learning the return distribution in a policy evaluation setting. The problem is set within the CliffWalk domain (Fig. \ref{fig:cliffwalk}). The transition dynamics follow those in \citet{rl:1998:sutton_barto:introduction_to_reinforcement_learning}. However, we include a five-percent chance of falling off the cliff from adjacent states. We used fixed Monte Carlo (MC) targets from the optimal greedy policy.

 Figure \ref{fig:cliffwalk} shows the convergence of the proximal loss and the mean square value error from the start state. As we can see, the estimated distribution (the histogram) accurately captures the target's features: the near certainty of walking off the cliff when moving right, the added chance of doing the same when choosing left or down, and finally the most profitable choice, moving up. 

\subsection{\textsc{wgf} in the Control Setting}
\begin{figure}
	\centering
	\includegraphics[width=\columnwidth]{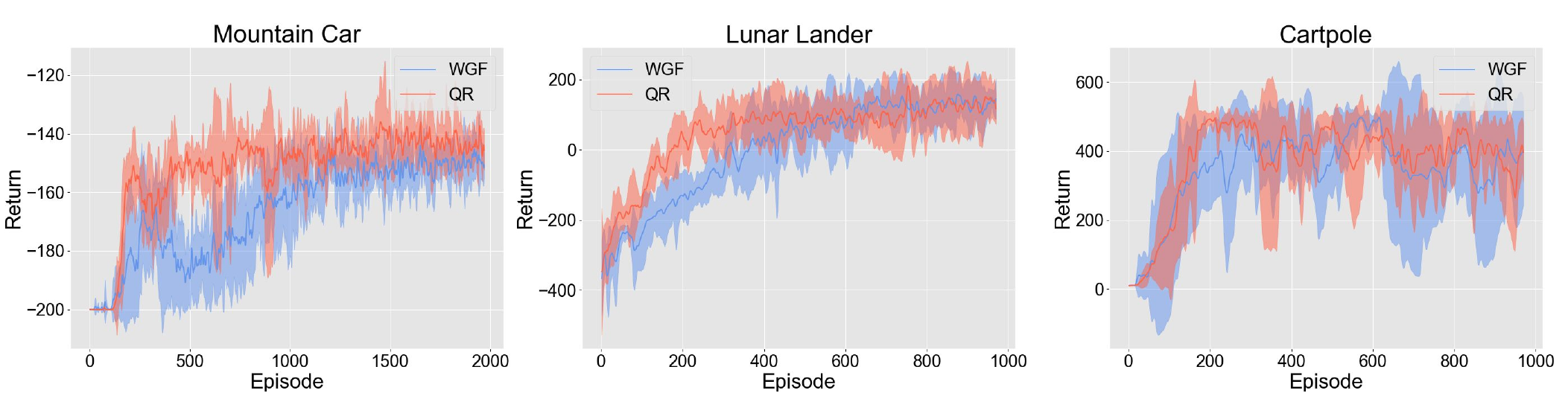}
	\caption{\textbf{Performance on control problems:} The \textsc{wgf} method matches the final average return of quantile regression.}
	\label{fig:func_approx}
\end{figure}

In this experiment we test the hypothesis that \textsc{wgf} Fitted $Q$-iteration is scalable to function approximation in the control setting. We parameterize return distributions with a two-layer fully-connected neural network of 256 hidden units. We use off-policy updates with bootstrapped targets and compare performance results with an agent trained using the \textsc{qr} loss \cite{distrl:2017:dabney_etal:distributional_reinforcement_learning_with_quantile_regression} on three common control tasks from the OpenAI Gym \cite{env:openaigym}: MountainCar, CartPole, and LunarLander. The results in Figure \ref{fig:func_approx} show that the \textsc{wgf} method matches the performance of \textsc{qr}.

\subsection{Control in the Presence of Uncertainty} 
This experiment studies how aleatoric uncertainty is handled during training. Specifically, we compare different policies for selecting among a multiplicity of competing solutions. We consider the $\varepsilon$-greedy, \textsc{ssd}, and $\textnormal{CVaR}_\alpha$ behavior policies for $\alpha\in\{0.05,0.25,0.45\}$. Each policy gathers data to update a greedy target policy. Different data distributions arise from the way each measures uncertainty. 

We expect the data distribution under the \textsc{ssd} policy to favor outcomes with higher certainty, because \textsc{ssd} compares the expected outcome over all represented risk levels. CVaR policies consider the expected outcome for a single risk level. Uncertainty drives action selection only when the specified risk level captures the true risk in the current state. Otherwise, we expect CVaR policies to become risk neutral.

 We revisit the CliffWalk domain with a modified reward structure (See appendix). Traversing the top and bottom rows have equal value. Each path has different reward uncertainty; the top row is deterministic, whereas the bottom row samples rewards from the Gaussian $\Ncal(-1,10^{-3})$. Under these conditions, we expect the \textsc{ssd} policy to prefer the top path and risk neutral methods to prefer the bottom row, since it will be more likely under a risk neutral policy.
 
\begin{figure}
	\centering
	\includegraphics[width=\columnwidth]{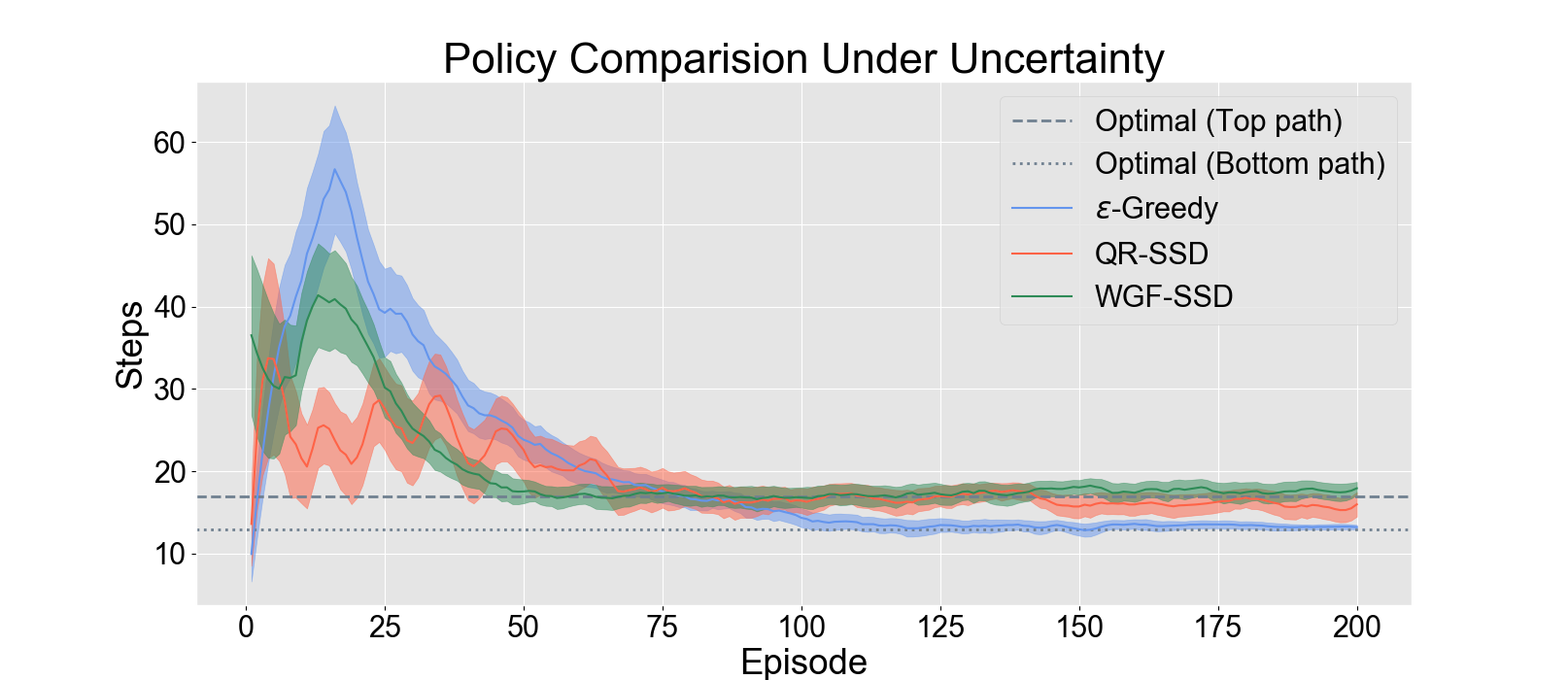}
	\includegraphics[width=\columnwidth]{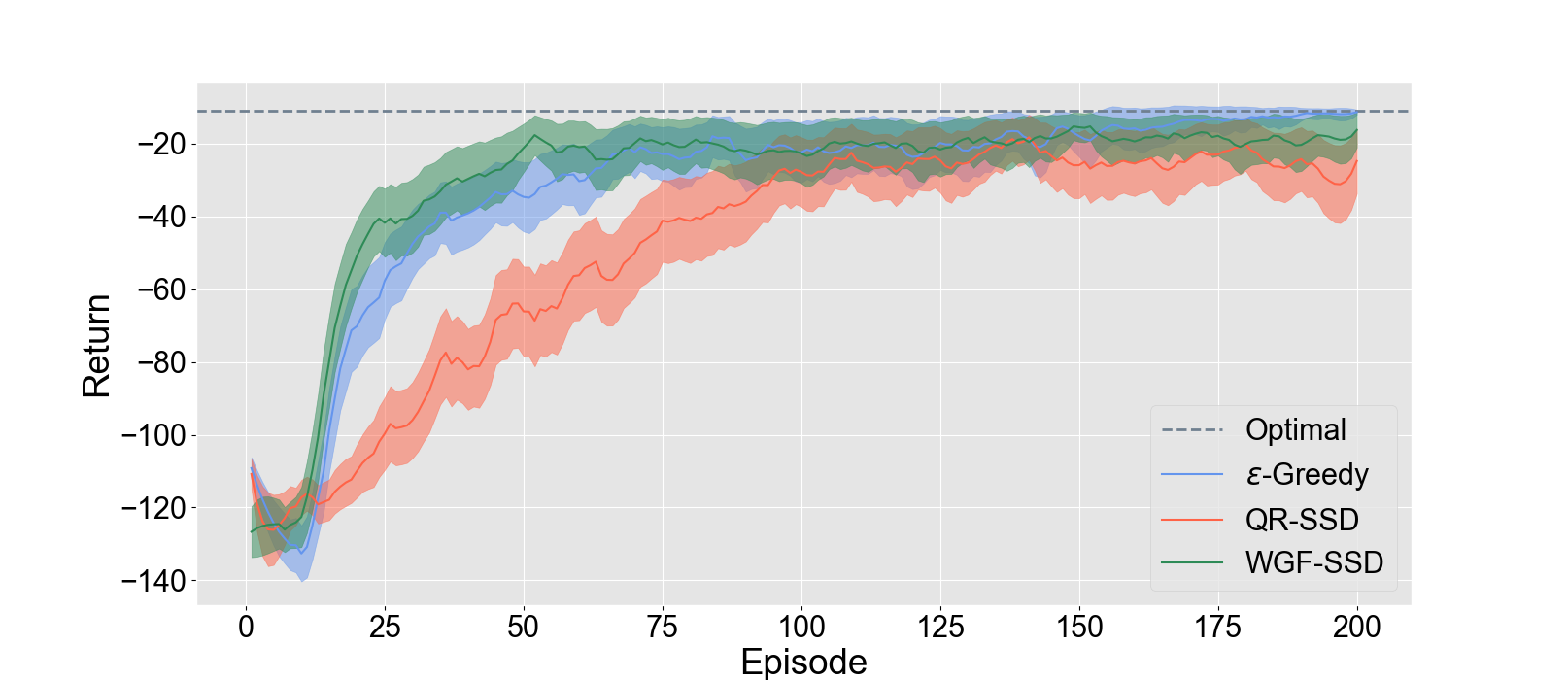}
	\caption{\textbf{The \textsc{ssd} behavior policy recovers the optimal target policy using samples from the least-disperse data distribution:} We compare the episodic step count and return using the \textsc{ssd} and $\varepsilon$-greedy policy. The distributional methods differ in their sample complexity but realize the same final solution.}
	\label{fig:mod_cliffwalk}
\end{figure}

 Figure \ref{fig:mod_cliffwalk} shows the average episodic step count and return, along with their 95\% confidence intervals computed from 50 trials. The step count data confirms our hypothesis that the \textsc{ssd} policy induces the least-disperse data distribution, since it takes the top path on average. We can also confirm that the $\varepsilon$-greedy policy chooses the bottom path, which is more likely under the sampling distribution from $s_0$ and incidentally more dispersed. We observe similar behavior between \textsc{qr} to \textsc{wgf} $Q$ iteration, consistent with results in Figure \ref{fig:regression}. Both methods induce similar data distributions over the top path at around the 75th episode. And in this domain, the \textsc{wgf} method learns the quickest. 

\begin{figure}
	\vspace{-.5em}
	\centering
	\includegraphics[width=\columnwidth]{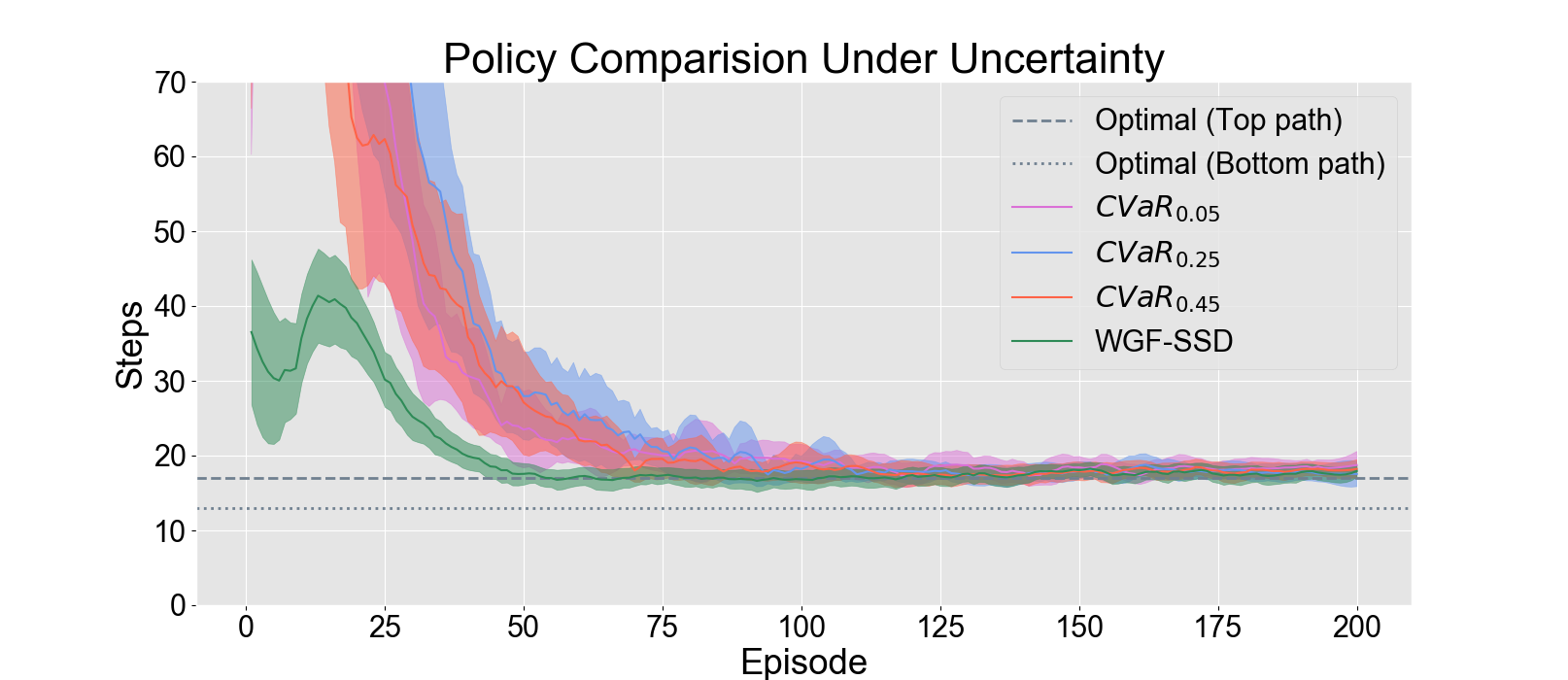}
	\includegraphics[width=\columnwidth]{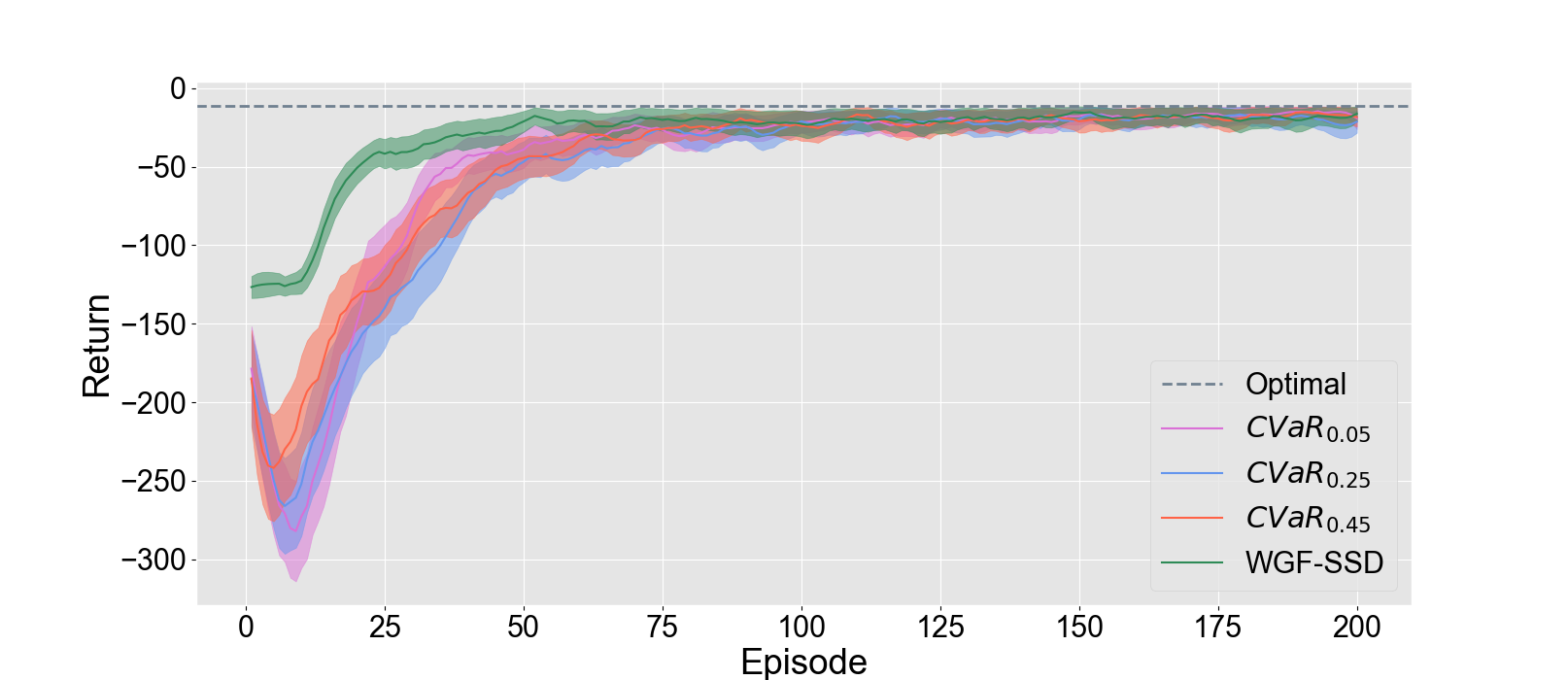}
	\includegraphics[width=\columnwidth]{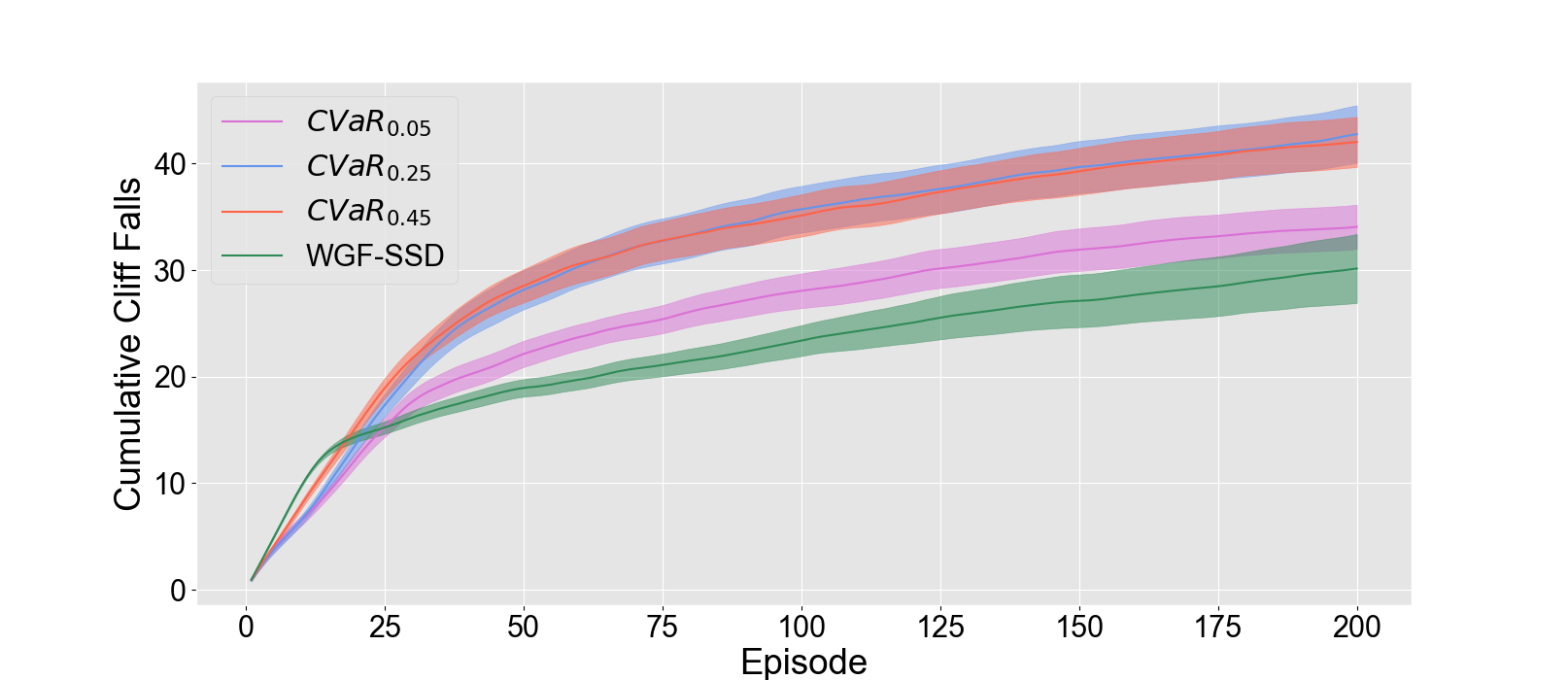}
	\caption{\textbf{Using many risk levels can improve exploration:} One risk level is not always appropriate for every state. Here, the CVaR policy leads the agent away from its goal, causing it to explore more than with the \textsc{ssd} policy, which uses many risk levels.}
	\label{fig:cvar_cliffwalk}
		\vspace{-3.5em}
\end{figure}

We find the greatest differences between the \textsc{ssd} and CVaR policies occur in the transient phase of learning (Figure \ref{fig:cvar_cliffwalk}). The CVaR agent takes more exploratory steps as a result of using a single uniform risk level. In high-stakes settings, the consequence of exploration can vary from undesirable to catastrophic. Here a cliff fall models a very costly outcome. Figure \ref{fig:cvar_cliffwalk} shows the number of cliff falls encountered throughout learning. Using the \textsc{ssd} policy results in a significantly lower number of these experiences. We interpret this as positive evidence to suggest that \textsc{ssd} provides a more comprehensive measure of uncertainty than CVaR. 


\section{Conclusion}
This paper argues for the use of \textsc{ssd} to select among a multiplicity of competing solutions. This can be useful in settings where one wishes to minimize exposure to uncertainty. We presented a convergent, online algorithm for learning return distributions (\textsc{wgf} Fitted $Q$-iteration). Our simulations demonstrated the algorithm can learn good policies, and that it scales up to function approximation. Based on our experimental results, we concluded that the \textsc{ssd} behavior policy can reduce dispersion in the data distribution and improve exploration in the presence of uncertainty.   

\section*{Acknowledgements}
The authors wish to thank the anonymous reviewers for their feedback. A special thanks goes to Marc G. Bellemare and to Shruti Mishra for their thoughtful reviews of earlier drafts. 

This work relates to Department of Navy award N00014-20-1-2570 issued by the Office of Naval Research. The United States Government has a royalty-free license throughout the world in all copyrightable material contained herein. This work was also supported in part by the National Science Foundation, grant number IIS-1652064, by the Robert Crooks Stanley Graduate Fellowship in Engineering and Science, and by the U.S. Department of Homeland Security under Cooperative Agreement No. 2014-ST-061-ML0001. The views and conclusions contained in this document are those of the authors and should not be interpreted as necessarily representing the official policies, either expressed or implied, of the U.S. Department of Homeland Security.

\bibliography{ref}
\bibliographystyle{icml2020}

\clearpage
\appendix

\newcommand{\myapptitle}{Stochastically Dominant Distributional Reinforcement Learning Appendix}
\icmltitlerunning{\myapptitle}

\onecolumn
\icmltitle{\myapptitle}




\icmlkeywords{Machine Learning, ICML}




\section{Experimental Details}

\subsection{Regression Comparison}
This experiment compared empirical first and second moment estimates between quantile regression and solutions of a Wasserstein gradient flow. The distributions were parameterized with the same number of particles, which we varied for values of 5, 10, 20, and 50. Particles were trained on data from a five-component Gaussian mixture model of those sample sizes. We draw samples from each component with equal probability $c_i=1/5$, using the means $\mu_i\in\{-5, -3, 0, 5, 6, 9\}$, and standard deviations $\sigma_i\in \{1, 2, 1, 2, 1, 0.5\}$, for $i=1,\cdots,5$. Models were evaluated on a separate draw of the same size as the training set. We computed the target values, $y$, using empirical estimates from $10^4$ samples. The violin plots show the distribution of root mean square error $RMSE=\sqrt{\frac{1}{N}\sum_{n=1}^N(y-\hat{y})^2}$ samples between the targets and the estimates $\hat{y}$ over $N=100$ trials.

\subsection{Ablation Study}
\begin{figure}[H]
	\centering
	\includegraphics[width=\columnwidth]{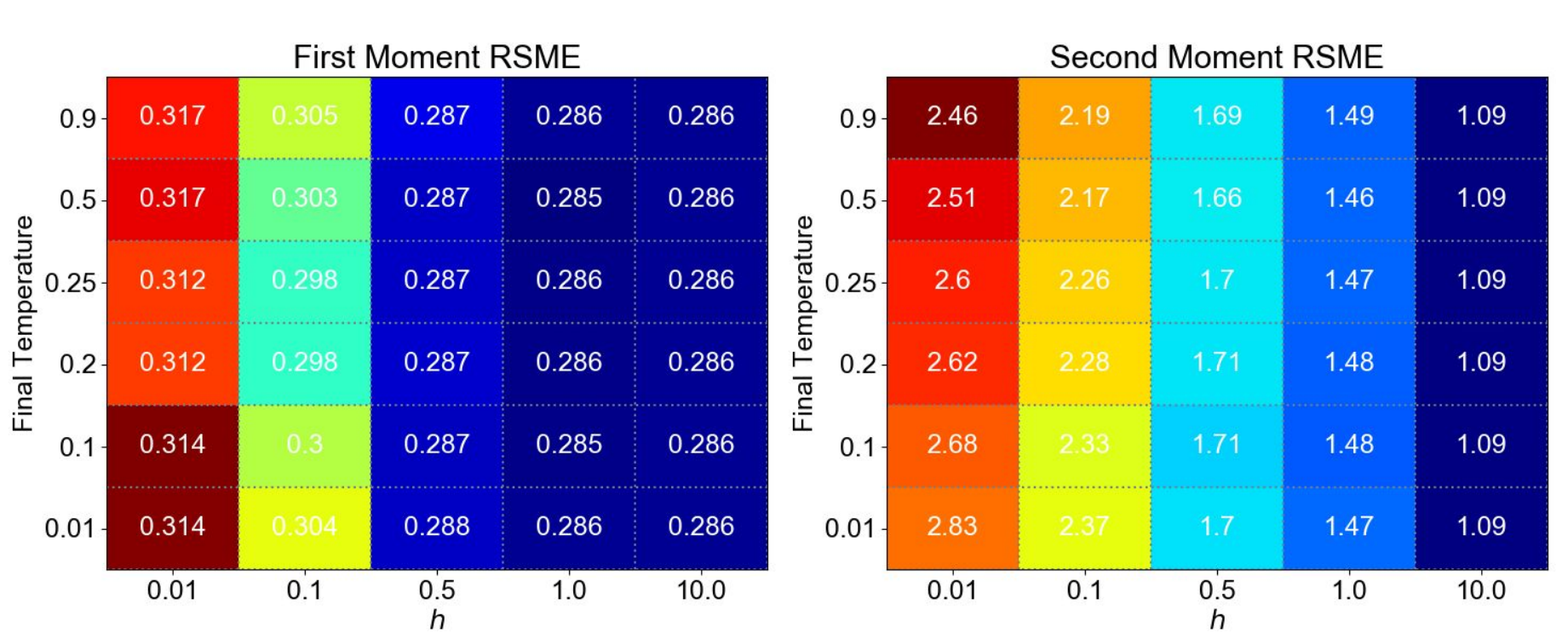}
	\caption{\textbf{Ablations:} Temperature $\beta$ and step size $h$ in proximal loss.}
	\label{fig:ablation}
\end{figure}

We ablated the minimum temperature $\beta^{-1} \in \{0.01, 0.1, 0.2, 0.25, 0.5, 0.9\}$ and step size $h \in \{0.01, 0.1, 0.5, 1., 10.\}$ over 50 trials. Data came from the five-component mixture model used in the Regression Comparison experiment. We report the root mean square error in the first and second moments with targets computed using $10^4$ samples and empirical estimators.

\subsection{\textsc{wgf} Policy Evaluation}
Here we perform policy evaluation on Monte Carlo (MC) returns from the optimal policy. The optimal policy was obtained by running $Q$-learning for $10^4$ episodes with an $(\epsilon =0.1)$-greedy behavior policy, $\gamma =0.9$, learning rate $\alpha=0.5$, and using an absorbing terminal state. MC returns were computed for each state from 200 rollouts of 200 time steps. We parameterized a discrete distribution with 200 particles initialized from a standard $\Ncal(0,1)$ Gaussian, then transported them using 100 gradient steps with a step size of $0.5$. The proximal loss was annealed down from $\beta^{-1}=1$ to $0.25$ in minimum steps of $0.5$; the proximal time step was set to $h=1$. We report the curves of the proximal loss and the squared value error at each gradient step.    

\subsection{\textsc{wgf} in the Control Setting}
This experiment used the OpenAI Gym \cite{env:openaigym} environments MountainCar, CartPole, and LunarLander with discrete actions. We estimated particle locations using a two layer fully-connected neural network, each with 256 hidden units. We trained these networks with the \textsc{wgf} proximal loss and the quantile regression loss from \cite{distrl:2017:dabney_etal:distributional_reinforcement_learning_with_quantile_regression}. Both models regressed 2 quantiles. We used the Adam optimizer \cite{opt:2015:kingma:adam} with a step size of $10^{-3}$. We used experience replay with batches of size 32 and a total capacity of $10^4$. Agents explored with an $(\epsilon=0.1)$-greedy policy, using $\gamma=0.99$ until the absorbing state was reached. We report data for 5 independent trials.

\subsection{Control in the Presence of Uncertainty}
\begin{figure}[H]
	\centering
	\includegraphics[width=0.8\columnwidth]{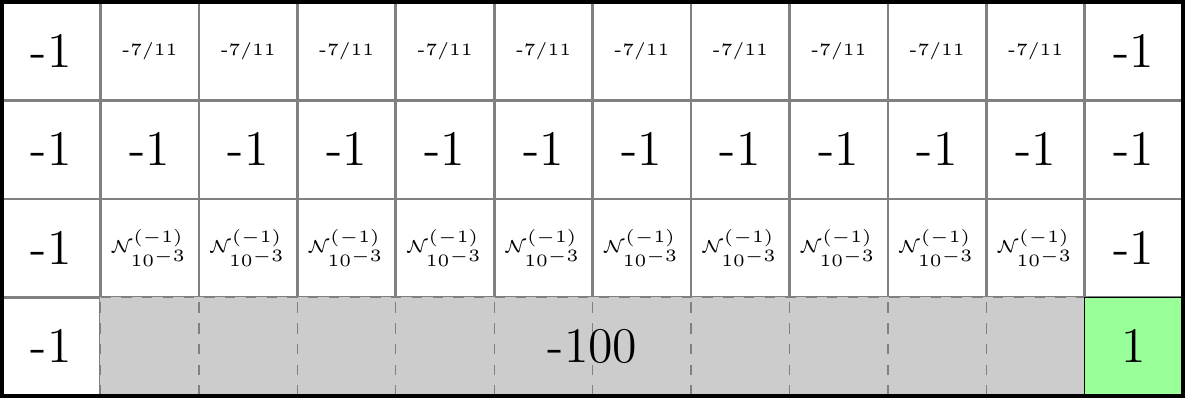}
	\caption{Modified CliffWorld with multiple solutions.}
	\label{fig:mod_cliff_walk}
\end{figure}
This experiment used data generated in the CliffWalk environment \cite{rl:1998:sutton_barto:introduction_to_reinforcement_learning}. The agent moves in four cardinal directions. We modified the reward function so they were assigned randomly according to Figure \ref{fig:mod_cliff_walk}. Here $\Ncal^{(\mu)}_\sigma$ denotes $\Ncal(\mu,\sigma)$. For the random rewards, we clipped them to be within the interval $[-10,10]$. Both the \textsc{wgf} and quantile regression agents used a tabular representation of 16 particles for each return distribution. Learning occurred with $\gamma=1$, a horizon length of 500, and the same loss settings used in the policy evaluation experiment. However, the number of gradient steps was limited to 50, unless a tolerance of $10^{-8}$ was exceeded below first. We report data gathered from $M=50$ independent trials. The 95\% confidence intervals were computed using the standard $t$-distribution with $M-1$ degrees of freedom.

\section{Mathematical Proofs and References to Supporting Results}
This section provides proofs to our main theoretical results. For our supporting results, we provide references to their original sources. We drop the superscript notation introduced in the main paper, used to denote single measures for state-action pairs. All the following results involving probability measure apply for single measures.

\setcounter{lemma}{0}
\setcounter{proposition}{0}
\setcounter{theorem}{0}
\begin{lemma}\label{lem:ssd}
	Let $\tau \in (0,1)$ and consider $\xi_\tau = F^{-1}_X(\tau)$. Then $F^{-2}_X(\tau) =\Ebf[X\leq \xi_\tau]$.
	\begin{proof}
		By conjugate duality, 
		\begin{align*}
			F^{-2}_X(\tau) &= \tau \xi_\tau - F^{(2)}_X(x),\\
			 &= \tau \xi_\tau - \tau\Ebf[X-\xi_\tau|X\leq \xi_\tau],\\
			 &=\tau\Ebf[X|X\leq \xi_\tau],\\
			 &=\Ebf[X\leq \xi_\tau].
		\end{align*}
	\end{proof}
\end{lemma}

\begin{proposition}
$Z^{(s,a_1)} \succeq_{(2)} Z^{(s,a_2)}$ if, and only if $\sum_{i=1}^jz^{[i]}_{a_1} \geq \sum_{i=1}^jz^{[i]}_{a_2}, \ \forall \ j=1,\cdots,N.$
\begin{proof}
We prove the result in the context of random returns. However, this holds for general random variables. We consider two random returns induced by the actions $a_1$ and $a_2$, respectively denoted $Z^{(s,a_1)}$, $Z^{(s,a_2)}$. Each return is approximated with a discrete Lagrangian measure 
\begin{align*}
	\mu^{(s,a_1)}&\approx\frac{1}{N}\sum_{n=1}^N\delta_{z^{(n)}_{a_1}}, & \mu^{(s,a_s)}&\approx\frac{1}{N}\sum_{n=1}^N\delta_{z^{(n)}_{a_2}}.
\end{align*}
Given that $Z^{(s,a_1)} \succeq_{(2)} Z^{(s,a_2)}$, we know by the definition that $F^{-2}_{Z^{(s,a_1)}}(\tau) \geq F^{-2}_{Z^{(s,a_2)}}(\tau)$ for all $\tau \in (0,1)$. Invoking Lemma \ref{lem:ssd} allows us to rewrite the definition with total expectations
\begin{align*}
		\Ebf[Z^{(s,a_1)} \leq \xi^{(\tau)}_{a_1}] &\geq \Ebf[Z^{(s,a_2)}\leq \xi^{(\tau)}_{a_2}], \ \forall \ \tau \in (0,1).
\end{align*}
Denote the ordered coordinates of a return distribution to be $z^{[1]} \leq z^{[2]}\leq \cdots \leq z^{[N]}$. Then with particle sets from each measure, we have
\begin{align*}
	\sum_{i=1}^jz^{[i]}_{a_1} &\geq \sum_{i=1}^jz^{[i]}_{a_2}, \ \forall \ j=1,\cdots,N.
\end{align*}
The other implication follows by normalizing the sums with $1/N$ and invoking Lemma \ref{lem:ssd} again to arrive at the definition.
\end{proof}
\end{proposition}

\begin{proposition}[\citet{sdom:1980:fishburn:stochastic_dominance_and_moments_of_distributions}]\label{lemma:moment_ordering}
	Assume $\mu$ has two finite moments. Then $X\succeq_{(2)} Y$ implies $\mu_X^{(1)} \geq \mu_Y^{(1)} $ or $\mu_X^{(1)} = \mu_Y^{(1)}$ and $\mu_X^{(2)} \leq \mu_Y^{(2)} $, where $(\cdot)$ denotes a particular moment of the distribution $\mu$.   
	\begin{proof}
		This result follows from Theorem 1 of \citet{sdom:1980:fishburn:stochastic_dominance_and_moments_of_distributions}, which proves an ordering dominance of any finite degree.  
	\end{proof}
\end{proposition}
	
\begin{proposition}
	Let $\{\mu_t\}_{t\in[0,1]}$ be an absolutely-continuous curve in $\Pscr(\mathbb{R})$ with finite second-order moment. Then for $t\in[0,1]$, the vector field $\vbf_t = \nabla(\frac{\delta E}{\delta t}(\mu))$ defines a gradient flow on $\Pscr(\mathbb{R})$ as $\partial_t \mu_t =- \nabla \cdot(\mu_t\vbf_t)$, where $\nabla\cdot \ubf$ is the divergence of some vector $\ubf$.
	\begin{proof}
		See \citet{gflow:2005:ambrosio:gradient_flows_in_metric_spaces_and_in_the_space_of_probability_measures}, Theorem 8.3.1.
	\end{proof}
\end{proposition}


\begin{proposition}\label{lemma:ito_conv}
	Let $\mu_0\in\Pscr_2(\mathbb{R})$ have finite free energy $E(\mu_0)<\infty$, and for a given $h>0$, let $\{\mu^{(h)}_t\}_{t=0}^K$ be the solution of the discrete-time variational problem, with measures restricted to $\Pscr_2(\mathbb{R})$, the space with finite second moments. Then as $h\rightarrow 0$, $\mu_K^{(h)} \rightarrow \mu_T$, where $\mu_T$ is the unique solution of the Fokker-Plank equation at $T=hK$.
	\begin{proof}
		See \citet{gflow:1998:jordan_etal:the_variational_formulation_of_the_fokker_planck_equation}, Theorem 5.1.
	\end{proof}
\end{proposition}

\begin{proposition}
	Let $\{\mu^{(h)}_t\}_{t=0}^K$ be the solution of the discrete-time JKO variational problem, with measures restricted to $\Pscr_2(\mathbb{R})$, the space with finite second moments. Then $E(\mu_t)$ is a decreasing function of time. 
	\begin{proof}
		We show that the free-energy $E(\mu) = F(\mu)+\beta^{-1} H(\mu)$ is a Lyapunov functional for the Fokker-Planck (FP) equation. Following the approach of \cite{gflow:1999:markowich_villani:on_the_trend_to_equilibrium_for_the_fokkerplanck_equation}, we consider the change of variables $\mu_t = h_te^{- U}$, where we let $\beta=1$ without loss of generality. With this, FP is equivalent to 
		\begin{align}\label{eq:fp_h}
			\partial_t h_t &= \Delta h_t -\nabla U \cdot \nabla h_t.
		\end{align}	
		Whenever $\phi$ is a convex function, one can check the following is a Lyapunov functional for \eqref{eq:fp_h}, and equivalently FP:
		\begin{align*}
			\int \phi(h_t)e^{- U}dz = \int \phi(\mu_t e^{U})e^{- U}dz .
		\end{align*}
		Differentiating with respect to time shows
		\begin{align*}
			\frac{d}{dt}\int \phi(h_t)e^{- U}dz = -\int \phi''(h_t)|\nabla h_t|^2e^{- U}dz  < 0.
		\end{align*}
		Now consider $\phi(h_t) = h_t\log(h_t)-h_t+1$. With the identity $\int(h_t-1)e^{- U}dz=0$, we find
		\begin{align*}
			\int \phi(h_t)e^{- U}dz &= \int \mu_t \log\left(\frac{\mu_t}{e^{- U}}\right)dz = \int \mu_t( U+\log\mu_t) dz = E(\mu).
		\end{align*}
		Thus, the free-energy functional is a Lyapunov function for the Fokker-Planck equation, and $E(\mu_t)$ is a decreasing function of time. In the low-energy state the optimal distributional Bellman equation is satisfied with pure Brownian motion.
	\end{proof}
\end{proposition}

\begin{theorem}
If $\Tcal\mu=\mu$, then $\textnormal{Prox}_{hE}^{\Wscr}(\mu)=\mu$ as $\beta \rightarrow \infty$.	\begin{proof}
		Let $d(\mu,\nu)$ be some distributional distance between measures $\mu$ and $\nu$, such as the supremal $k$-Wasserstein $= \sup_{s,a}\Wscr_k(\mu,\nu)$. Furthermore, suppose $\mu^* = \Tcal\mu^*$ is the fixed point of the optimal distributional Bellman operator $\Tcal$. We consider the proximal operator 
		\begin{align*}
			\text{Prox}_{hE}^{\Wscr}(\mu_k) = \argmin_{\mu} \Wscr^2_2(\mu,\mu_k)+2h E(\mu).		
		\end{align*}
	It follows that $\mu^* = \Tcal\mu^*$ and 
		\begin{align*}
				d(\Tcal\mu^*,\mu^*) &\leq d(\text{Prox}_{hE}^{\Wscr}(\mu^*), \mu^*) = d\biggl(\argmin_{\mu} \Wscr^2_2(\mu,\mu^*)+2h \underbrace{E(\mu)}_{0 \text{ as } \beta \rightarrow \infty},\mu^*\biggr),\\
				&\leq d\biggl(\argmin_{\mu} \Wscr^2_2(\mu,\mu^*)=\mu^*,\mu^*\biggr)\\
				&\leq 0
		\end{align*} 
		Distance is non-negative, so it must be that $\text{Prox}_{hE}^{\Wscr}(\mu^*) = \Tcal\mu^* = \mu^*$.
	\end{proof}
\end{theorem}




\section{Expanded Background}
\subsection{Euclidean Gradient Flows}
Suppose we have a smooth function $F\colon\mathbb{R}^d\rightarrow\mathbb{R}$ and an initial point $\xbf_0\in\mathbb{R}^d$. The gradient flow of $F(\xbf)$ is defined as the solution to the differential equation $\frac{d\xbf}{d\tau}=-\nabla F(\xbf(\tau))$, $\tau >0$, and $\xbf(0)=\xbf_0$. This has a unique solution if $\nabla F$ is Lipschitz continuous.

Exact solutions are typically intractable. A standard numerical method, called the Minimizing Movement Scheme (MMS) \cite{gflow:1999:gobbino:minimizing_movements_and_evolution_problems_in_euclidean_spaces}, evolves $\xbf$ iteratively for small steps along the gradient of $F$ at the current point $\xbf_k$. The next point is 
\begin{align*}
	\xbf_{k+1} &= \xbf_k -\nabla F(\xbf_{k+1})h,
\end{align*} 
for the step size $h$. Determining $\xbf_{k+1}$ is equivalent to solving the optimization problem
\begin{align*}
	\xbf_{k+1} \in \argmin_{\xbf} F(\xbf) +\frac{||\xbf-\xbf_k||_2^2}{2h}.
\end{align*}
Where the squared Euclidean norm is denoted $||\cdot||_2^2$. Convergence of the sequence $\{\xbf_k\}$ to the exact solution has been established for this method, \cite{gflow:2005:ambrosio:gradient_flows_in_metric_spaces_and_in_the_space_of_probability_measures}.

\subsection{Sinkhorn's Algorithm}
We describe how the Kantorovich problem can be made tractable through entropy regularization, then present an algorithm for approximating the $\Wscr_2^2$ distance. The key message is that including entropy reduces the original Optimal Transport problem to one of matrix scaling. Sinkhorn's algorithm can be applied for this purpose to admit unique solutions. 

The optimal value of the Kantorovich problem is the exact $\Wscr_2^2$ distance. Given probability measures $\alpha = \sum_{i=1}^N\alpha_i\delta_{x_i}$ and $\beta = \sum_{j=1}^M\beta_j\delta_{y_j}$, the problem is to compute a minimum-cost mapping, $\pi$, defined as a non-negative matrix on the product space of atoms $\{x_1,\cdots,x_N\}\times\{y_1,\cdots,y_M\}$. Denoting the cost to move $x_i$ to $y_j$ as $C_{ij} = ||x_i-y_j||^2$, we have
\begin{align}\label{eq:kantorovich}
	\Wscr_2^2(\alpha,\beta) &= \min_{\pi\in \mathbb{R}^{N\times M}_{\geq 0}}\left< \pi , C \right> =\sum_{ij}\pi_{ij}C_{ij},\\
	& \text{such that } \pi \mathbf{1}_M = \alpha, \  \pi^\top \mathbf{1}_N = \beta. 
\end{align}
This approach constitutes a linear program, which unfortunately scales cubically in the number of atoms. We can reduce the complexity by considering an entropically regularized version of the problem. Let $\varepsilon$ be a regularization parameter. The new problem is written in terms of the generalized Kullback Leibler (KL) divergence:
\begin{align}\label{eq:entropic_kantorovich}
	\Wscr_2^2(\alpha,\beta)\approx \Wscr_\varepsilon(\alpha,\beta) &= \min_{\pi\in \mathbb{R}_{\geq 0}^{N\times M}} \left< \pi, C\right>  + \varepsilon \mathsf{KL}(\pi || \alpha \otimes \beta ),\\
	 &= \sum_{i,j} \pi_{ij}C_{ij} + \varepsilon\sum_{i,j} [\pi_{ij}\log\frac{\pi_{ij}}{\alpha_i\beta_j} -\pi_{ij} + \alpha_i\beta_j ], \\
	& \text{such that } \pi \mathbf{1}_M = \alpha, \  \pi^\top \mathbf{1}_N = \beta. 
\end{align}
The value of $\Wscr_\varepsilon(\alpha,\beta)$ occurs necessarily at the critical point of the constrained objective function
\begin{align}
	L_\varepsilon = \sum_{i,j}\pi_{ij}C_{ij} &+ \varepsilon\sum_{i,j} [\pi_{ij}\log\frac{\pi_{ij}}{\alpha_i\beta_j} -\pi_{ij} + \alpha_i\beta_j ]\nonumber\\
	 &- \sum_{i}f_i\biggl(\sum_j\pi_{ij} - \alpha_i\biggr) - \sum_{j}g_j\biggl(\sum_i\pi_{ij} - \beta_j\biggr),\\
	\frac{\partial L_\varepsilon}{\partial \pi_{ij}}&= 0 \Longrightarrow \ \forall \ i,j,\  C_{ij}+ \varepsilon\log \frac{\pi_{ij}^*}{\alpha_i\beta_j} = f_i^* +g_j^*.\label{eq:opt_cond}
\end{align}
The last line of \eqref{eq:opt_cond} shows that the entropically-regularized solution is characterized by two vectors $f^*\in \mathbb{R}^N$, $g^*\in \mathbb{R}^M$. With the following definitions
\begin{align}\label{eq:sinkhorn_vars}
	u_i &= \exp(f_i^*/\varepsilon), & v_j &= \exp(g_j^*/\varepsilon), & K_{ij} = \exp(-C_{ij}/\varepsilon),
\end{align}
we can write the optimal transport plan as $\pi^* = \mathbf{diag}(\alpha_iu_i)K\mathbf{diag}(v_j\beta_j)$. And the approximate Wasserstein distance can be computed simply as
\begin{align*}
	\Wscr_\varepsilon(\alpha,\beta) &= \left<\pi^*,C\right> + \varepsilon\mathsf{KL}(\pi^*||\alpha\otimes\beta) = \sum_{ij}(f^*_i + g_j^*) = \left<f^*,\alpha\right> + \left< g^*, \beta\right>
\end{align*}
We mentioned that Optimal Transport reduces to positive matrix scaling. Indeed, using the vectors $u$ and $v$, Sinkhorn's algorithm provides a way to iteratively scale $K$ such that the unique solution is $\pi^*$. Initialize $u^{(0)}=\mathbf{1}_N$, and $v^{(0)}=\mathbf{1}_M$, then perform the following iterations for all $i,j$
\begin{align}
	v^{(1)}_j &= \frac{1}{[K^\top(\alpha \odot u^{(0)})]_j}, & u^{(1)}_i &= \frac{1}{[K(\beta \odot v^{(1)})]_i},\nonumber\\
	\vdots &  & \vdots &\nonumber\\
	v^{(n+1)}_j &= \frac{1}{[K^\top(\alpha \odot u^{(n)})]_j}, & u^{(n+1)}_i &= \frac{1}{[K(\beta \odot v^{(n+1)})]_i}.\label{eq:sinkhorn_iter}
\end{align}
Sinkhorn's algorithm performs coordinate ascent with $f$ and $g$ to maximize the dual maximization problem
\begin{align}\label{eq:dual_entropic_kantorovich}
	\Wscr_\varepsilon(\alpha,\beta) &= \max_{f\in\mathbb{R}^N,g\in\mathbb{R}^M} \left<f,\alpha\right> + \left< g, \beta\right> - \varepsilon \left< \alpha \otimes \beta , \exp\{ (f \oplus g - C )/\varepsilon\}-1\right>.
\end{align}
Each update consists of kernel products, $K^\top(\alpha \odot u)$ and $K(\beta \odot v)$, and point-wise divisions. We describe this procedure in Algorithm \ref{alg:sinkhorn}, using computations in the log domain to numerically stabilize the updates. The log updates derive from \eqref{eq:sinkhorn_vars} and \eqref{eq:sinkhorn_iter}:
\begin{align}
 	\log v_j &= -\log\sum_{i}K_{ij}\alpha_iu_i & \log u_i &= -\log\sum_j K_{ij}\beta_jv_j,\nonumber \\
 	g_j &= -\varepsilon\log\sum_i\exp\{ (-C_{ij} + f_i)/\varepsilon + \log \alpha_i \} & 
 	f_i &= -\varepsilon\log\sum_j\exp\{ (-C_{ij} + g_j)/\varepsilon + \log \beta_j\}.\label{eq:sinkhorn_iter}
\end{align}
The Sinkhorn iterations typically loop until convergence. In practice, we choose a decreasing temperature sequence $\{\varepsilon_n\}$ with which to bound the number of iterations.

\begin{algorithm}[H]
	\caption{Sinkhorn's Algorithm in the log domain for $\Wscr_2^2$}
	\label{alg:sinkhorn}
	\begin{algorithmic}[1]
	\STATE \textbf{input:} Source and target measures $\alpha = \sum_{i=1}^N\alpha_i\delta_{x_i}$, $\beta = \sum_{j=1}^M\beta_j\delta_{y_j}$, Annealing temperature sequence $\{\varepsilon_n\}$
	\STATE {\color{gray} \# Initialize dual variables}
	\STATE $i \in \{1,\cdots,N\}$, $j \in \{1,\cdots,M\}$
	\STATE $f_i \gets 0$, $g_j \gets 0$ $\forall$ $i,j$
	\STATE {\color{gray} \# Perform coordinate ascent in the log domain}
	\FOR{$\varepsilon \in \{\varepsilon_n\} $}
		\STATE $C_{ij} = \frac{1}{2\varepsilon}||x_i-y_j||^2$ $\forall$ $i,j$
		\STATE $g^{(n+1)}_j \gets -\varepsilon \log \sum_{i} \exp\{(-C_{ij} + f^{(n)}_i)/\varepsilon + \log \alpha_i\}$ $\forall$ $j$
		\STATE $f^{(n+1)}_i \gets -\varepsilon \log \sum_{j} \exp\{(-C_{ij} +  g^{(n+1)}_j)/\varepsilon +\log \beta_j\}$ $\forall$ $i$
	\ENDFOR
	\STATE {\color{gray} \# Return the entropic-regularized OT distance}
	\STATE \textbf{output:} $\left< f,\alpha\right> + \left< g,\beta\right>$
	\end{algorithmic}
\end{algorithm}

\section{Supporting Results}
\subsection{Proof that the Gibbs measure minimizes free energy.}
\begin{remark}
	Let $E(\mu)= F(\mu) +\beta^{-1}H(\mu)$, with $F(\mu) = \int U(z)d\mu$. The minimizer is the Gibbs density, 
	\begin{align*}
		\mu_*(z) &= Z^{-1}\exp\{-\beta\psi(z)\},
	\end{align*} 
	where $\psi(z) = U(z)+\int_0^1\lambda(\tau)S(z,\tau)d\tau$, and $Z= \int\exp\{-\beta\psi(z)\}dz$.

	\begin{proof}
		We set the functional derivative, or the first variation, of $E$ to zero and solve for $\mu$. The derivatives are
		\begin{align*}
			\frac{\delta F}{\delta \mu} &= U(z), & \frac{\delta H}{\delta \mu} &= \log(\mu) + 1.
		\end{align*}
		Solving for $\mu_*$ emits a proportionality, which can be normalized as described:
		\begin{align*}
			U(z) +\beta^{-1}(\log(\mu_*) + 1) = 0 \implies
			\mu_* \propto \exp\{-\beta\psi(z)\}
		\end{align*}
	\end{proof}
\end{remark}

\subsection{On the prevalence of multiple solutions}
\begin{figure}[H]
	\centering
	\includegraphics[width=0.45\columnwidth]{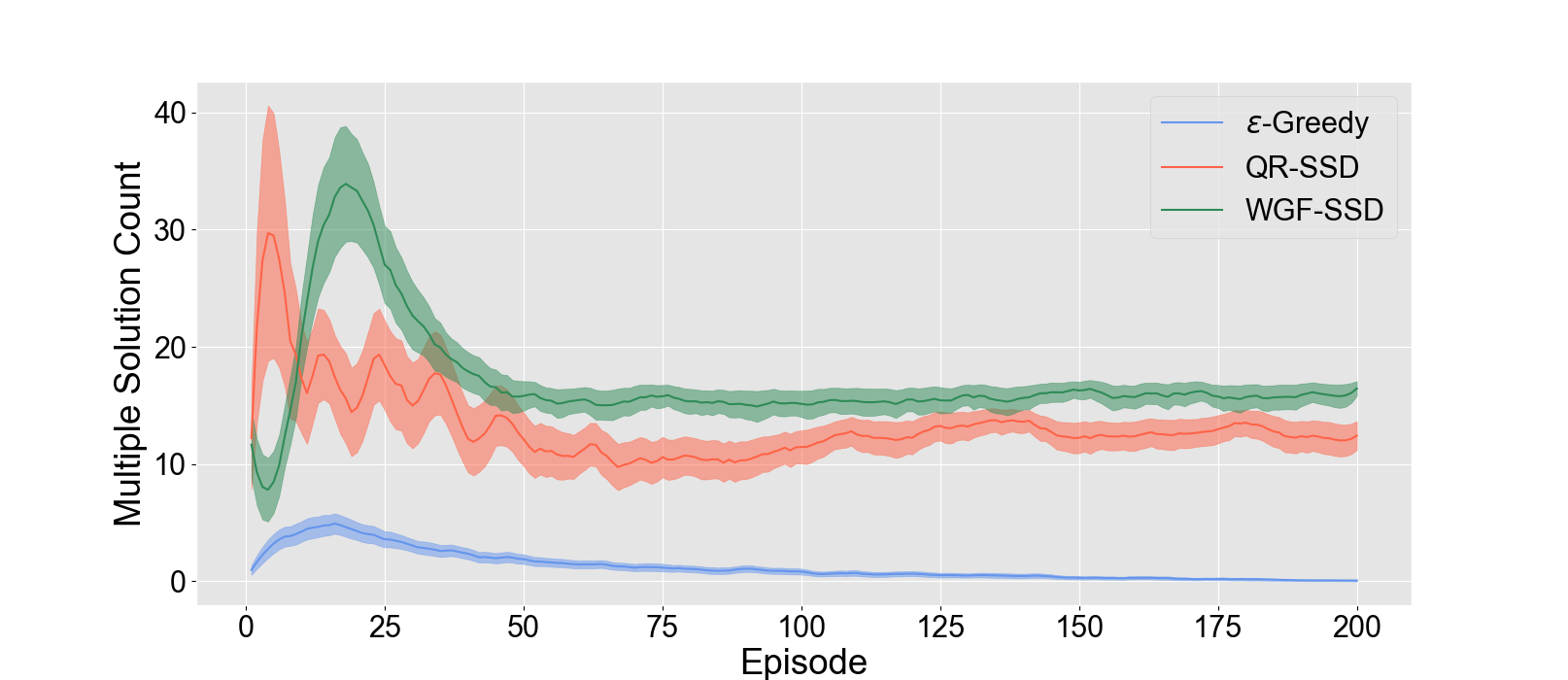}
	\includegraphics[width=0.45\columnwidth]{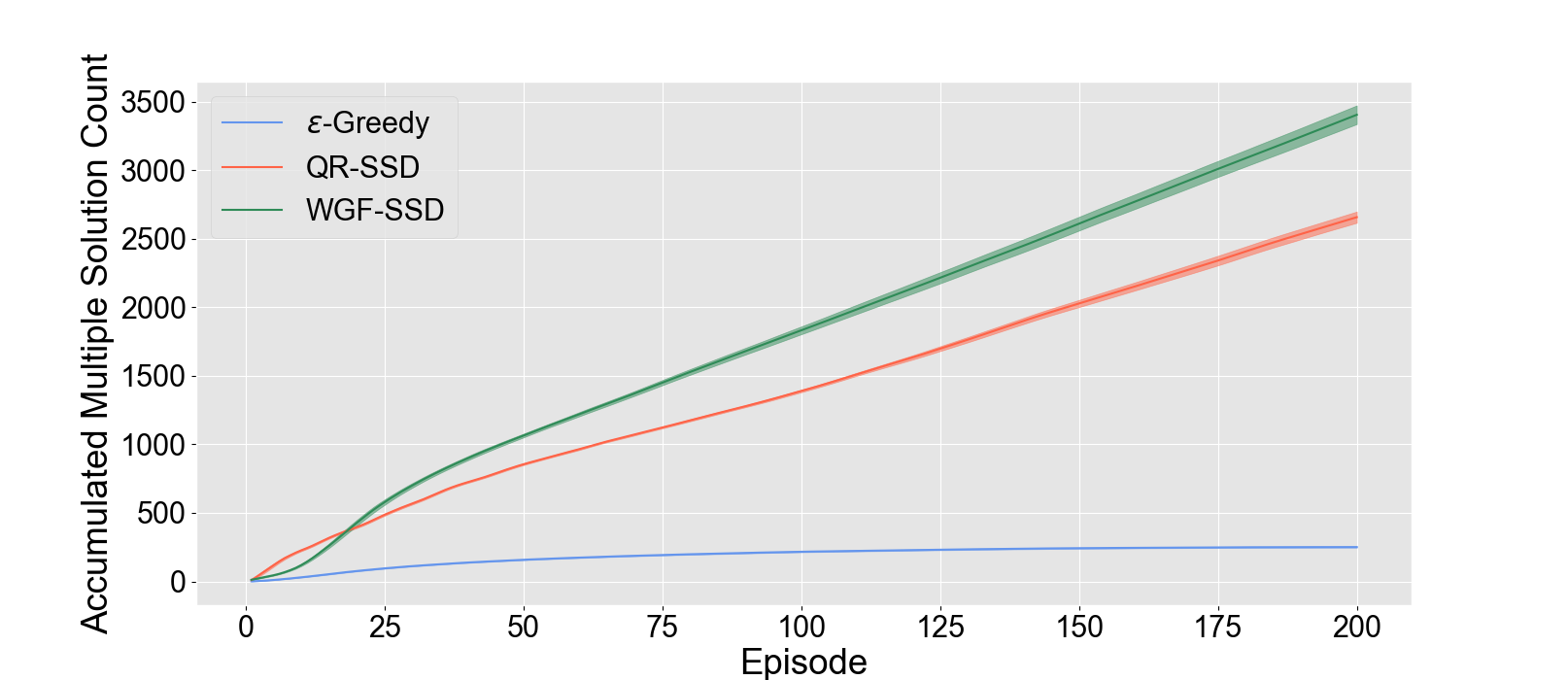}
	\includegraphics[width=0.45\columnwidth]{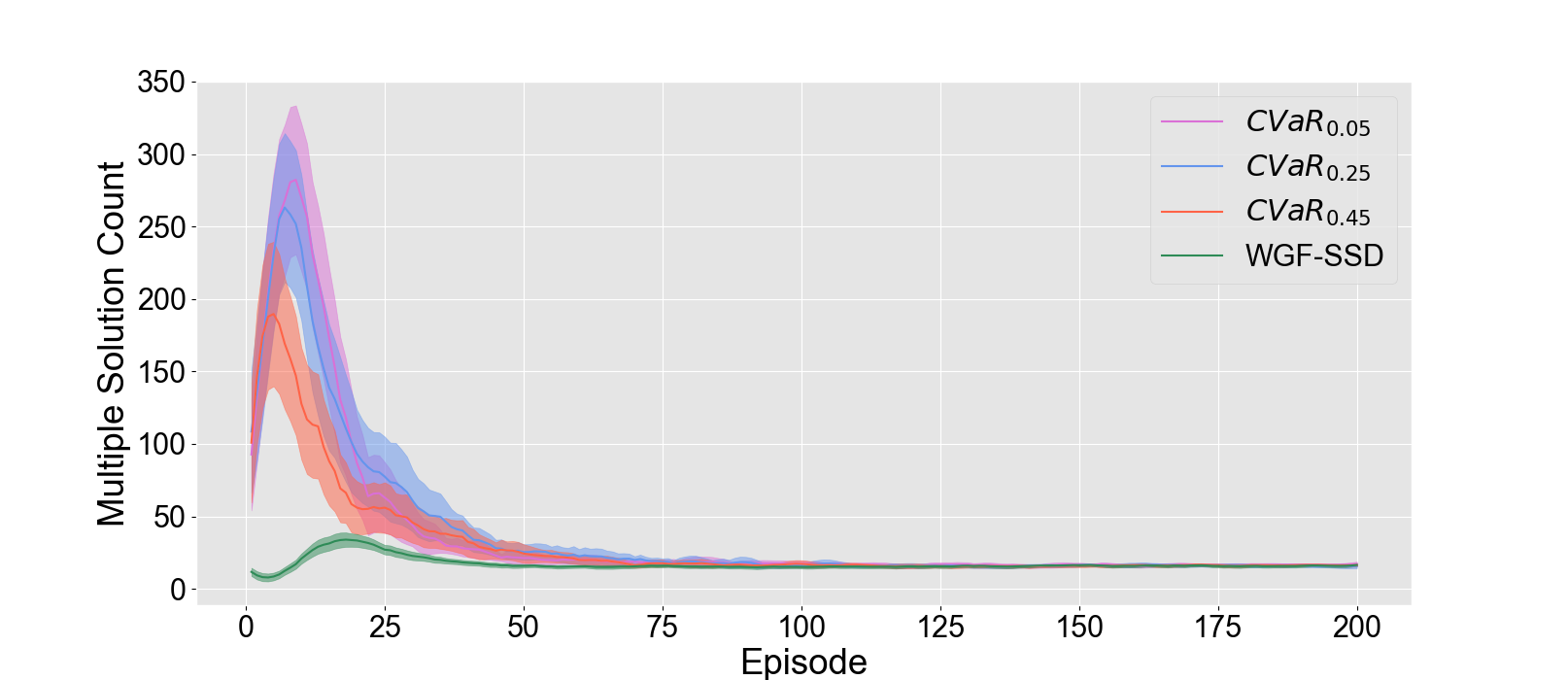}
	\includegraphics[width=0.45\columnwidth]{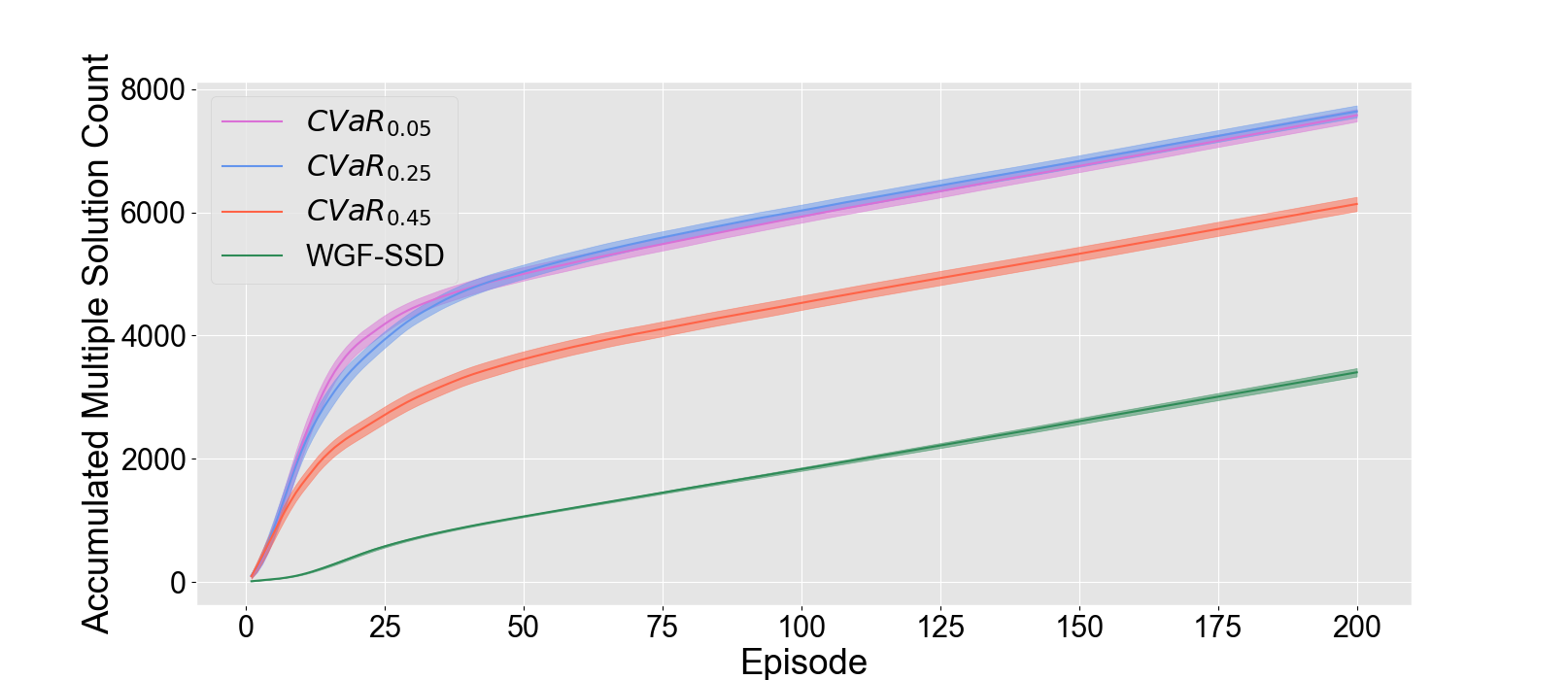}
	\caption{Frequency of multiple-solution events that occurred during the Control in the Presence of Uncertainty experiment.}
	\label{fig:multiple_sols}
\end{figure}

We are concerned with settings where the agent must select between multiple equivalently-valued actions in a way that minimizes uncertainty. Figure \ref{fig:multiple_sols} shows the number of times these events occurred during the Control in the Presence of Uncertainty experiment. We present this data to support the claim that multiple solutions occur often enough in our experiment to merit a policy for selecting among the options. 

\bibliography{ref}
\bibliographystyle{icml2020}

\end{document}